\newif\ifsuppl \suppltrue  

\newif\ifsmallfonts \smallfontstrue

\newif\ifcitenum \citenumfalse

\documentclass{article}

\usepackage{microtype}
\usepackage{booktabs} 

\usepackage{silence}
\usepackage{hyperref}

\usepackage[accepted]{icml2020}

\setcitestyle{open={[},close={]},numbers}

\usepackage{enumitem}
\usepackage{placeins}
\usepackage{subcaption}  
\usepackage{xspace}
\usepackage{booktabs} 
\usepackage{amsmath}
\usepackage[most]{tcolorbox}
\usepackage{tcolorbox}
\usepackage[T1]{fontenc}
\usepackage{amsfonts,amssymb}
\usepackage{bigdelim}
\usepackage{tabularx}
\usepackage{bbm}
\usepackage{calc}
\usepackage{footmisc}
\usepackage{mathrsfs}
\usepackage{mathtools}
\usepackage{stfloats}  
\usepackage{setspace}
\usepackage{tikz}
\usepackage{textgreek}
\usepackage{graphicx}
\usetikzlibrary{tikzmark}
\usetikzlibrary{arrows.meta}
\usetikzlibrary{bending}
\usetikzlibrary{calc,positioning,quotes}
\usetikzlibrary{matrix,shapes.arrows}

\renewcommand{\algorithmicrequire}{\textbf{Input:}}
\renewcommand{\algorithmicensure}{\textbf{Output:}}

\newcommand{\citeinline}[1]{\cite{#1}}  

\newcommand{\algcomment}[1]{{\color{gray}\# #1}}
\newcommand{\algspacedcomment}[1]{{\color{gray}\,\,\# #1}}
\newcommand{\alghlight}[1]{{\color{blue}#1}}
\newcommand{\alghlightcolor}{blue}

\makeatletter
\DeclareRobustCommand\onedot{\futurelet\@let@token\@onedot}
\def\@onedot{\ifx\@let@token.\else.\null\fi\xspace}
\def\eg{\textit{e.g}\onedot}
\def\Eg{\textit{E.g}\onedot}

\def\vs{\textit{vs}\onedot}

\makeatother

\newcommand{\supplsection}[3]{
    \ifsuppl%
        \section{#1}
        \label{#2}
        #3
    \else%
        \par\refstepcounter{section}%
        \sectionmark{#1}%
        \label{#2}
    \fi%
}

\newcommand{\supplheader}{
    \renewcommand{\figurename}{Supplementary Figure}
    \renewcommand{\tablename}{Supplementary Table}
    \renewcommand{\thesection}{S\arabic{section}}
    \renewcommand{\thefigure}{S\arabic{figure}}
    \renewcommand{\thetable}{S\arabic{table}}
    \setcounter{section}{0}
    \setcounter{figure}{0}
    \setcounter{table}{0}
    \ifsuppl%
        \FloatBarrier
        \clearpage
        \twocolumn[
        \icmltitle{\ourtitle}
        \centerline{\textbf{\LARGE Supplementary Material}}
        \vspace{30pt}
        ]
    \fi%
}

\makeatletter
\renewcommand*{\ALG@name}{Method}
\makeatother

\def\codefont{\fontfamily{lmtt}\selectfont}
\newcommand{\textcode}[1]{{\normalfont\codefont #1}}

\newcounter{codelinecounter}[section]
\definecolor{codebackground}{rgb}{1.0,1.0,1.0}
\definecolor{codebackgroundprogress}{rgb}{1.0,0.99,0.98}
\definecolor{codeframe}{rgb}{0.8,0.8,0.8}
\definecolor{codegreen}{rgb}{0,0.4,0}
\definecolor{codeblue}{rgb}{0.25,0.25,0.75}
\definecolor{codegray}{rgb}{0.5,0.5,0.5}
\def\codefontsize{\fontsize{8.5}{9}\selectfont}  
\newcommand{\codeline}[1]{{#1\par}}
\newcommand{\codelinenum}{\stepcounter{codelinecounter} {\color{codegray} \ifnum\value{codelinecounter}<10 0\fi\arabic{codelinecounter}}\ }

\newcommand{\codecomment}[1]{{\color{codegray} \# #1}}

\newcommand{\codedef}[1]{{\color{codegreen} #1}}

\newcommand{\codetab}{~~}
\newcommand{\codeskip}{\smallskip}
\newenvironment{code}[3]{  
    \setcounter{codelinecounter}{0}
    \begin{tcolorbox}[
        width=#1,height=#2,
        valign=center,left=0pt,right=0pt,top=0pt,bottom=0pt,
        colback=#3,colframe=codeframe,boxrule=0.5pt,arc=0pt]
    \codefont
    \codefontsize
}{ 
    \end{tcolorbox}
}

\newcommand{\experimentdetails}[2][Experiment Details]{
    \begin{spacing}{0.85}%
    {\ifsmallfonts\footnotesize\fi\rule{20pt}{0.4pt}\ \textit{#1: #2}%
    \hrulefill
    }%
    \vspace{-\lineskip}%
    \end{spacing}%
    \vspace{\parskip}%
}

\usepackage{todonotes}

\newif\iftodos \todostrue

\makeatletter
\newcommand\thefontsize{{[[[This font is \f@size pt]]]}}
\makeatother

\usepackage[english]{babel}
\usepackage{blindtext}
\usepackage{lipsum}

\def\ourtitle{Unified Functional Hashing in Automatic Machine Learning}

\icmltitlerunning{Unified Functional Hashing}

\begin{document}

\twocolumn[
\icmltitle{\ourtitle}

\icmlsetsymbol{equal}{*}
\icmlsetsymbol{senior}{\textdagger}

\begin{icmlauthorlist}
\icmlauthor{Ryan Gillard}{research}
\icmlauthor{\ \ \ \ Stephen Jonany}{exgoogle}
\icmlauthor{\ \ \ \ Yingjie Miao}{research}
\icmlauthor{\ \ \ \ Michael Munn}{research}
\icmlauthor{\ \ \ \ Connal de Souza}{google}
\icmlauthor{Jonathan Dungay}{exgoogle}
\icmlauthor{\ \ \ \ \ \ \ Chen Liang}{research}
\icmlauthor{\ \ \ \ \ \ \ David R. So}{research}
\icmlauthor{\ \ \ \ \ \ \ Quoc V. Le}{research}
\icmlauthor{\ \ \ \ \ \ \ \ Esteban Real}{research}
\end{icmlauthorlist}

\icmlaffiliation{research}{Google Research, Mountain View, CA, USA}
\icmlaffiliation{google}{Google, Mountain View, CA, USA}
\icmlaffiliation{exgoogle}{For work done while at Google, Mountain View, CA, USA}

\icmlcorrespondingauthor{Esteban Real}{ereal@\allowbreak google.com}

\icmlkeywords{machine learning, neural networks, evolution, evolutionary computation, evolutionary algorithms, regularized evolution, reinforcement learning, program synthesis, architecture search, NAS, neural architecture search, neuro-architecture search, AutoML, AutoML-Zero, algorithm search, meta-learning, genetic algorithms, genetic programming, neuroevolution, neuro-evolution, autorl, auto-rl, functional equivalence, functional equivalence cache, FEC}

\vskip 0.3in
]

\printAffiliationsAndNotice{}  

\begin{abstract}

The field of Automatic Machine Learning (\mbox{AutoML}) has recently attained impressive results, including the discovery of state-of-the-art machine learning solutions, such as neural image classifiers. This is often done by applying an evolutionary search method, which samples multiple candidate solutions from a large space and evaluates the quality of each candidate through a long training process. As a result, the search tends to be slow. In this paper, we show that large efficiency gains can be obtained by employing a fast \emph{unified functional hash}, especially through the \emph{functional equivalence caching} technique, which we also present. The central idea is to detect by hashing when the search method produces equivalent candidates, which occurs very frequently, and this way avoid their costly re-evaluation. Our hash is ``functional'' in that it identifies equivalent candidates even if they were represented or coded differently, and it is ``unified'' in that the same algorithm can hash arbitrary representations---\eg compute graphs, imperative code, or lambda functions. As evidence, we show dramatic improvements on multiple \mbox{AutoML} domains, including neural architecture search and algorithm discovery. Finally, we consider the effect of hash collisions, evaluation noise, and search distribution through empirical analysis. Altogether, we hope this paper may serve as a guide to hashing techniques in \mbox{AutoML}.


\end{abstract}

\section{Introduction}
\label{intro_sec}


The last decade has seen dramatic growth in the complexity of machine learning methods. Image classifiers, for example, have reached hundreds of layers in depth, with intricate paths from input to output. Their discovery required years of dedication by multiple human experts, who do not have enough time to optimize models for all the possible applications that require them. This has fueled the resurgence of \emph{Automatic Machine Learning (AutoML)}, a field that aims to improve a given aspect of an ML system without human participation. An example is \emph{neural architecture search}, where the structure of an image classifier, seen as a compute graph, is optimized for accuracy \cite{baker2016designing,zoph2016neural,tan2019mnasnet}. The AutoML paradigm involves first defining a large search space with candidate solutions (\eg the space of compute graphs) and then searching for the best one. One way to perform this search is through evolutionary computation, which has been a traditional approach in AutoML \cite{miller1989designing,angeline1994evolutionary,stanley2002neat} and is still used in modern studies, largely due to its simplicity and ability to reach state-of-the-art results (\eg \citeinline{real2018regularized}). Like many of the alternatives, evolutionary computation requires evaluating multiple candidates, rendering the search process slow, as each evaluation involves training an ML model. This paper demonstrates, however, that the efficiency of evolutionary AutoML can be greatly improved through the use of a \emph{unified functional hash} (UFH) that allows us to quickly determine the equivalence of candidates without full training. This equivalence detection, in turn, permits us to reduce the number of costly evaluations significantly, as evolution tends to produce many equivalent candidates whose evaluation can be skipped. Importantly, our method detects equivalence in function and not just structure. In biological terms and loosely speaking, functional hashing allows us to compare ``phenotypes'' instead of ``genotypes''. We will show that new and classic evolutionary techniques can employ this hash to improve the outcomes of the search process. While we will focus on evolutionary computation because of its long AutoML tradition, we highlight that the UFH can also be applied to other popular search methods, such as reinforcement learning (Section~\ref{discussion_robustness_sec}).

Fast functional equivalence detection helps the search process because evolutionary computation often produces equivalent candidates with high frequency within an experiment. If a candidate is detected to be equivalent to a previously seen candidate, for example, its evaluation result can be retrieved from a cache, saving compute time. An obvious possible approach for detecting the equivalence of two candidates is to compare their structures. In the case of neural architectures represented as compute graphs, this involves comparing the graphs. This approach has the following drawbacks, as illustrated by Figure~\ref{functional_structural_fig}. First, finding such isomorphisms is likely not solvable in polynomial time. Second, graphs may still perform the same computation even if they are not isomorphic. For example, two graphs that differ by a multiplication by the constant 1 are structurally different but functionally equivalent. Our functional equivalence method does not suffer from these drawbacks: as we will see, computing the functional hash scales well with the graph size and the resulting hash is independent of the structure. More generally, the hash computation is, by construction, much faster than the evaluation of the ML model. The hash is only sensitive to what we ultimately care about, the function being represented, regardless of structure. Moreover, any method that compares structures must be modified for the different representations used in different domains. In architecture search, for example, we are concerned with directed acyclic graphs, while in program discovery we are concerned with segments of code that read and write into a virtual memory. The two representations are structurally very different from each other, so comparing \emph{structure} will require different methods. On the other hand, as we explain next, comparing \emph{function} can be done with one single method---UFH.

We propose to quickly hash candidates with our UFH method,\footnote{\label{conference_publication_footnote}A reduced version of this idea is tangentially mentioned in our group's earlier conference paper \citeinline{real2020automl}; see Section~\ref{related_work_sec} for details.} as follows. In all setups under consideration, candidates encode a transformation from input to output. For example, an image classifier takes an input image represented as a matrix of floating-point numbers and outputs a list of logits. To compute the hash, we first construct a sequence of canonical inputs, \eg by picking a fixed sequence of images. Feeding those inputs into the model results in a corresponding sequence of outputs that can be interpreted as a fingerprint of the model. We will refer to these as \emph{hashable outputs}. We then convert the hashable outputs to a single integer hash with a hash-mixing process. If enough inputs are used, this hash becomes a good fingerprint. As it turns out, the number of inputs required to effectively hash the candidate is very small compared to the number of inputs required to train it, resulting in important compute reduction form the point of view of AutoML. In Section~\ref{methods_sec}, we describe the method in more detail, including how to handle non-determinism in the evaluation; in Section~\ref{methods_hashing_sec}, we explain how to determine a sufficient number of canonical inputs; and in Section~\ref{hash_collisions_sec}, we present a method for handling occasional collisions.

To demonstrate the usefulness of UFH, we use it to implement three evolutionary techniques and demonstrate the resulting gains in various AutoML setups. The first of the techniques is the \emph{functional equivalence cache (FEC)} of evaluation results, which we also present in this paper.\footnotemark[\getrefnumber{conference_publication_footnote}] The second technique is the \emph{functional change mutation} (FCM), which prevents insertion of children that are identical to their parents \cite{singh2006comparison}. The third technique is the use of a \emph{tabulist} that prevents the insertion into the population of individuals that have been recently seen \cite{glover1989tabu,glover1995genetic,yuen2008genetic}. Each of these permits skipping a costly evaluation when the functional equivalence of two models has been detected. Each of the techniques has their own strengths: FEC guarantees unaltered evolutionary dynamics under certain conditions, FCM is convenient as hashes do not need to be stored, and tabulist permits the escape from local optima. Section~\ref{methods_techniques_sec} will describe these techniques in more detail. We will show that FEC improves outcomes in all the setups we considered, while the other methods are either neutral or positive (Section~\ref{results_sec}).

We chose diverse experimental setups, representative of various styles of AutoML: MNAS aims to discover neural network architectures for image classification, AutoML-Zero evolves ML learning algorithms represented as code, and AutoRL evolves reinforcement learning loss functions represented as compute graphs. Further justification for using MNAS as a baseline is its state-of-the-art discoveries and the popularity of neural architecture search. Further justification for AutoML-Zero is its novelty recognized in a Humies award\footnote{\url{https://www.human-competitive.org/awards}}. Further justification for AutoRL is the demonstration that it can evolve algorithms that surpass classic results like DQN \cite{mnih2013dqn} in certain environments; moreover, the especially noisy nature of RL training makes it an appealing challenge for us. In addition to those three, we also evaluate on NAS-Bench-101, a popular AutoML benchmark that simulates a neural architecture search task through table lookups. This means that NAS-Bench-101 has a relatively small search space but incredible evaluation speeds, making it convenient for testing the limitations of our methods. Therefore, we use NAS-Bench-101 to show robustness to experimental conditions such as the search algorithm used (Section~\ref{discussion_robustness_sec}), and to study the impact of distributed search, hash collisions, and noise (Section~\ref{fec_guarantee_sec}).

In summary, our contributions are:
\begin{itemize}[noitemsep,topsep=-5pt,leftmargin=*,labelsep=4pt]
    \item A \emph{unified functional hash} (UFH) for ML algorithms, largely agnostic to their representation and structure;
    \item The technique of \emph{functional equivalence caching} (FEC) and the application of UFH to two classic techniques;
    \item A demonstration of the effectiveness of these techniques, spanning multiple search spaces; and
    \item Guidelines for how to use functional hashing, with examples of its limitations and benefits.
\end{itemize}
Our open-source code can be found online\footnote{\url{https://github.com/google-research/unified_functional_hashing}}.

\section{Related Work}
\label{related_work_sec}


Our work relates to the field of evolutionary AutoML, an intersection between automatic machine learning (\emph{AutoML}, \citeinline{hutter2019automated}) and evolutionary computation \cite{holland1992adaptation,mitchell1998introduction,jong2009evolutionary}. Evolutionary computation dates back decades (see history chapter in \citeinline{jong2009evolutionary}). Loosely speaking, it comprises the use of a population-based method inspired by biological evolution to find an optimum within a search space that maximizes a function called the \emph{fitness}. In the case of its application to AutoML, the search space is comprised of learning systems, such as neural networks. The points in the search space may differ, for example, in their architecture, in which case we refer to it as neural architecture search or \emph{NAS} \cite{stanley2002evolving,real2017large,elsken2019neural}. Other search spaces can involve only the hyperparameters of the learning system (hyper-parameter optimization or \emph{HPO}, \citeinline{bergstra2011algorithms,feurer2019hyperparameter}), a specific operation \cite{liu2020evolving}, an optimizer \cite{chalmers1991evolution,bengio1994use,co2021evolving}, or end-to-end code \cite{real2020automl}, among others.

Evolutionary AutoML is an example of \emph{trial-based} AutoML, in which each candidate is trained independently. Other flavors of trial-based AutoML replace the evolutionary method in the outer loop with alternative approaches, such as random search \cite{bergstra2012random}, Bayesian optimization \cite{snoek2012practical,feurer2015efficient,white2021bananas}, or reinforcement learning \cite{baker2016designing,zoph2016neural,tan2021efficientnetv2}. Because all of these preserve the training inner loop, it is possible to apply our methods to them as well (\eg in Section~\ref{discussion_robustness_sec}, but generally out of scope for this paper). Trial-based AutoML is popular because it has obtained high quality results, but it often requires large amounts of computation. The alternative is \emph{efficient AutoML}, a collection of techniques that trade a small loss in the quality of results for large reductions in search compute \cite{andrychowicz2016learning,liu2018darts,cai2018proxylessnas,xie2018snas,zela2019understanding}. Our work aims at reducing the gap between trial-based AutoML and efficient AutoML without sacrificing quality. Some branches of AutoML can also be made more efficient through predictor neural nets \cite{white2021bananas,shi2020bridging,wen2020neural}. In these studies, a graph neural network (GNN) learns how candidates perform and this is used to guide the ongoing search. The GNN limits the application to cases where search candidates are represented as graphs. Our method, in contrast, applies to any representation (Section~\ref{setups_sec}). Moreover, our method is simple in that it does not require any additional learning or difficult meta-parameter optimization. Furthermore, our method is orthogonal to predictor-based approaches in that it can be used in combination with them.



Two of the techniques through which we apply UFH were based on existing literature, namely \emph{functional change mutation} (FCM) and \emph{tabulist}. The exact relationship of our implementations to the original ones is complex, as these ideas appear in multiple papers as different variants, often entangled with additional features. We have distilled them to their simplest forms, while ensuring compatibility with UFH. FCM is based on \citeinline{singh2006comparison}, especially their section on ``modified clearing''. Namely, FCM involves mutating a child until it is different from its parent \cite{mauldin1984maintaining,singh2006comparison}. It is particularly easy to implement as it requires no additional state for the evolutionary algorithm. A tabulist is a list of individuals that are to be denied entry into the population because they are in an over-explored area of the search space \cite{glover1989tabu,glover1995genetic,yuen2008genetic}. The key difference between our work and previous uses of FCM and tabulist is that we use our functional hash instead of a structural comparison.

Functional hashing can be used to fingerprint any aspect of an ML algorithm, including the training method and data augmentation, but one important aspect is the inference model or neural network, as in the case of MNAS (Section~\ref{mnas_setup_sec}). This is related to the idea of characterizing neural networks from their input-output function, which has been common in neuroscience since \citeinline{hubel1962receptive}. Applications can be found in computer science too; \eg \cite{erhan2009difficulty} also use a canonical set of inputs to produce characteristic outputs, which then are dimensionally reduced to display the evolution of the artificial neural network's weights during training. Instead, we use the outputs in order to produce a hash of the algorithm, for a different purpose, and for arbitrary learning algorithms.

Our main contribution is a systematic study of functional hashing through the three said techniques in four experimental setups, to demonstrate its breath of applicability. Our group's previous paper used UFH but only through the single FEC technique and in one experimental setup \cite{real2020automl}. Importantly, it did not control for overfitting, thus preventing attributing the improvement in the final results to the use of FEC. After our said publication, \citeinline{alet2020meta} independently used a similar technique (without baselines) to filter out all duplicated candidates before the main search procedure. Finally, \citeinline{co2021evolving} used the technique but a demonstration of its effectiveness was outside their scope. In this paper, we perform side-by-side ablations in all cases, always controlling for overfitting by reevaluating results on data unseen by the search process. We also go beyond FEC into multiple techniques and show effectiveness in multiple experimental setups chosen a priori. Additionally, we go beyond previous work in discussing limitations, such as hash collisions and the effect of noise, and we recommend best practices.

\section{Methods}
\label{methods_sec}



The Unified Functional Hashing (UFH) method at the core of this paper applies specifically to AutoML, as it takes advantage of the nested optimization structure commonly used in this field. Thus, we start by reviewing the typical AutoML paradigm in Section~\ref{methods_automl_sec}. Then, Section~\ref{methods_hashing_sec} follows with a description of UFH, our main methodological contribution. Finally, we describe the three techniques through which we apply UFH in Section~\ref{methods_techniques_sec}.

\subsection{Automatic Machine Learning}
\label{methods_automl_sec}

AutoML aims to automatically discover or improve aspects of machine learning, such as neural network architectures, training algorithms, or data augmentation strategies. The aspect to improve is referred to as the \emph{task}. Given a task, AutoML requires first defining a \emph{search space} of all possible candidate solutions to the task. For example, if we are to discover neural network architectures, the search space could be that of all directed acyclic compute graphs representing the connectivity of the neural network layers. Such search spaces are typically too large to explore exhaustively. Thus, the goal of AutoML is to find a good candidate in that space by only sampling a limited number $N$ of points in the space.

In this paper, we focus on the popular AutoML paradigm of \emph{trial-based evolutionary search}. This approach consists of a two-level nested search, with an evolutionary controller in the outer loop to propose new \emph{candidates} and an \emph{evaluation} inner loop to determine how good the candidates are. The outer loop controller proposes new candidates through an evolutionary method. Evolutionary methods work by iteratively improving a collection of \emph{candidates} called the \emph{population}. To start the process, the population is filled with \emph{seed} candidates; in the image classifier example, seed candidates could be just randomly wired architectures. Once an initial population of candidates is available, the controller proposes new candidates in cycles. In a typical approach, each cycle consists of \emph{selecting} from the population a good candidate called the \emph{parent} and cloning it to produce a new candidate called the \emph{child}. The child is then altered with a localized random change called a \emph{mutation}, and is finally added to the population. Because the selection of parents is biased toward good algorithms, the population improves over time. After many such evolutionary cycles, the best candidate is selected as the solution. There are many variants of evolutionary algorithms, but in this paper we will concentrate on \emph{regularized evolution} (Method~\ref{automl_outer_alg}), which has been popular in recent AutoML.

\begin{algorithm}
\caption{Regularized Evolution}
\label{automl_outer_alg}
\small
\begin{algorithmic}
\REQUIRE A search space of candidates $S$.
\REQUIRE An ``Evaluate'' function from $S$ to fitness.
\ENSURE A high-fitness candidate.
\STATE $population \gets$ empty queue
\STATE $n = 0$ \algspacedcomment{Number of sampled candidates.}
\STATE \algcomment{P is the population size (a meta-parameter).}
\WHILE{$|population| < P$}
    \STATE $seed \gets$ RandomCandidate($S$)
    \STATE $seed.fitness \gets$ Evaluate($seed$)
    \STATE $population.enqueue(seed)$
    \STATE $n = n + 1$
\ENDWHILE
\STATE \algcomment{N is the total number of candidates (a meta-parameter).}
\WHILE{n < N}
    \STATE \algcomment{T is the tournament size (a meta-parameter).}
    \STATE $tournament \gets$ RandomSubset($population$, size=$T$)
    \STATE $parent \gets $ HighestFitnessCandidate($tournament$)
    \STATE $child \gets$ Mutate($parent$)
    \STATE $child.fitness \gets$ Evaluate($child$)
    \STATE $population.enqueue(child)$
    \STATE $population.dequeue()$ \algspacedcomment{Remove oldest.}
    \STATE $n = n + 1$
\ENDWHILE
\STATE \textbf{Return} HighestFitnessCandidate($population$)
\end{algorithmic}
\end{algorithm}

The candidates proposed by the outer loop controller are evaluated in an inner loop. Evaluation of a candidate results in a measure of their quality called the \emph{fitness}. In the image classifier example, this evaluation would entail the training and validation of the neural network architecture over a labeled image dataset such as ImageNet; the fitness would then be the accuracy obtained. More generally, the evaluation inner loop iterates first over training examples (to optimize the candidate's continuous parameters through learning---\eg neural network weights) and then iterates over validation examples (to assess the quality of the model with the optimized parameters). This is summarized in Method~\ref{automl_inner_alg}. The computed fitness is then fed back to the outer loop controller to improve future candidate construction. Thus, we have a two-level, nested search process with all the discrete components optimized by the outer loop, while the inner loop handles continuous parameters. The inner loop evaluation is typical of machine learning, while the outer loop paradigm in the previous paragraph is typical of evolutionary computation.

\begin{algorithm}
\caption{AutoML Evaluation}
\label{automl_inner_alg}
\small
\begin{algorithmic}
\REQUIRE A candidate (determines the InitializePass, ForwardPass, and BackwardPass functions).
\ENSURE The candidate's fitness.
\STATE \algcomment{The typical AutoML evaluation is standard ML optimization.}
\STATE InitializePass() \algspacedcomment{\Eg randomly initialize model's weights.}
\STATE \algcomment{We assume a mini-batch of size 1 for simplicity.}
\FOR{$example$ \textbf{in} $training\ data$}
    \STATE \algcomment{\Eg in classification, forward outputs = predictions.}
    \STATE $forward\ outputs$ = ForwardPass($example.inputs$)
    \STATE BackwardPass($forward\ outputs$, $example.labels$)
\ENDFOR
\STATE $error$ = 0
\FOR{$example$ \textbf{in} $validation\ data$}
    \STATE $forward\ outputs$ = ForwardPass($example.inputs$)
    \STATE $error$ += Error($forward\ outputs$, $example.labels$)
\ENDFOR
\STATE \textbf{Return} Fitness($error$)
\end{algorithmic}
\end{algorithm}

Because the evolutionary process can overfit the data, it is necessary to follow it with a reevaluation of the best evolved candidates on unseen data. This is sometimes known as the \emph{meta-validation} phase of AutoML, by analogy with the validation phase of regular ML, as opposed to the above \emph{meta-training} phase which is typically analogous to the training phase of regular ML.

\subsection{Unified Functional Hashing}
\label{methods_hashing_sec}

Our main contribution, unified functional hashing (UFH), is a process to characterize an arbitrary candidate with a hash that is computed by ``hacking'' into the inner-loop structure of AutoML, as described in this section. The resulting hash can then be used by the outer loop controller to make decisions, as will be shown in the rest of the paper. We now describe how such a hash is constructed.

Method~\ref{hash_alg} shows the hashing calculation. Essentially, it consists of carrying out the same steps as in the AutoML evaluation of Method~\ref{automl_inner_alg}, while harvesting some floating-point values produced by the forward pass. We will refer to these values as \emph{hashable outputs}. The hashable outputs can be any by-product of the forward pass that depends critically on the candidate. For example, in the case of image classifier search, the logits can be used, as any meaningful change to architecture of the image classifier will affect the logits. The floating-point hashable outputs are then combined into an integer hash by the \textcode{AddToHash} function detailed in Method~\ref{add_to_hash_alg}. Because these hashable outputs depend critically on the candidate, so does the hash. Even though evaluation and hashing have many commonalities, hashing is much faster because only a tiny subset of the training and validation examples are required (see \textcode{SmallCanonicalSubset} lines in Method~\ref{hash_alg}).

\begin{algorithm}
\caption{Unified Functional Hashing}
\label{hash_alg}
\small
\begin{algorithmic}
\REQUIRE A candidate (determines the InitializePass, ForwardPass, and BackwardPass functions).
\ENSURE The candidate's hash.
\STATE \algcomment{Lines where UFH differs from AutoML evaluation}
\STATE \algcomment{are highlighted in \alghlightcolor}.
\STATE \algcomment{Seeding in next like is scoped to this method only.}
\STATE \alghlight{SeedRandomNumberGenerator(FIXED\_CONSTANT)}
\STATE InitializePass()
\STATE \alghlight{$hash$ = 0}
\FOR{$example$ \textbf{in} \alghlight{SmallCanonicalSubset(}$training\ data$\alghlight{)}}
    \STATE $forward\ outputs$, $hashable\ outputs$ = 
    \STATE \,\,\,\, ForwardPass($example.inputs$)
    \STATE BackwardPass($forward\ outputs$, $example.labels$)
    \color{\alghlightcolor}
    \FOR{$hashable\ output$ \textbf{in} $hashable\ outputs$}
        \STATE $hash$ = AddToHash($hash$, $hashable\ output$)
    \ENDFOR
    \color{black}
\ENDFOR
\FOR{$example$ \textbf{in} \alghlight{SmallCanonicalSubset(}$validation\ data$\alghlight{)}}
    \STATE $forward\ outputs$, $hashable\ outputs$ = 
    \STATE \,\,\,\, ForwardPass($example.inputs$)
    \color{\alghlightcolor}
    \FOR{$hashable\ output$ \textbf{in} $hashable\ outputs$}
        \STATE $hash$ = AddToHash($hash$, $hashable\ output$)
    \ENDFOR
    \color{black}
\ENDFOR
\color{blue}
\STATE \textbf{Return} $hash$
\color{black}
\end{algorithmic}
\end{algorithm}

\begin{algorithm}
\caption{AddToHash}\label{algorithm:addtohash}
\label{add_to_hash_alg}
\begin{algorithmic}
\REQUIRE $hash$ (integer hash value)
\REQUIRE $hashable\ output$ (floating-point number) 
\ENSURE Updated $hash$, with $hashable\ output$ mixed in.
    \STATE $sgn$ $\gets$ Sign($hashable\ output$)
    \STATE $exp$ $\gets$ Exponent($hashable\ output$)
    \STATE $man$ $\gets$ Mantissa($hashable\ output$)
    \STATE \algcomment{Truncation keeps only $m\_bits$ (a meta-parameter).}
    \STATE $man$ $\gets$ Truncate($man$, $m\_bits$)
    \FOR{$val$ \textbf{in} \{sgn, exp, man\}}
        \STATE $hash \gets \textrm{HashMix}(hash, val)$
    \ENDFOR
\STATE \textbf{Return:} $hash$
\end{algorithmic}
\end{algorithm}

It is important that the hashable outputs contain information about all aspects of the candidate, which is why we harvest them using a few steps of training and validation. Note that harvesting hashable outputs only during validation would prevent the hash from capturing information about the gradients. Also, harvesting only during training can result in subtle problems in certain search spaces: for example, AutoML-Zero can evolve code that manifests only in validation (due to instructions that cancel each other while executing \textcode{ForwardPass} and \textcode{BackwardPass} in alternation). It is also necessary to use enough examples as using too few may not probe all aspects of the candidate's input--output map. For example, in architecture search, dead ReLUs could zero out their upstream neural circuit for any one example. In our setups, we found about 10 examples to be generally sufficient.

The examples chosen must be \emph{canonical}---the same sequence must be used each time we compute a hash. (This stands in contrast to the evaluation protocol, where it is desirable to randomize the order of examples). Only with canonical examples will the hashable outputs constitute a characterization of the function of the candidate. For the same reason, any random number generator used must be reset with a fixed seed before hashing each candidate (\textcode{SeedRandomNumberGenerator} line in Method~\ref{hash_alg}). This is unusual but critical. For example, ML models often have their weights initialized with small random numbers (during the \textcode{InitializePass} line in Methods~\ref{automl_inner_alg} and~\ref{hash_alg}). It is also common to shuffle examples or to randomize activations for regularization purposes, as in the dropout technique. All these involve random number generation. Resetting the generator at the beginning of hashing ensures that two identical candidates will have identical hashable outputs. This way, the hash will remain consistent in spite of the random number generation. Occasionally, we may get unlucky with the random seed of the generator and end up with a pathological initialization. To avoid this, we can pick a few fixed random seeds and repeat the hashable outputs harvesting with each of them (not included in Method~\ref{hash_alg} pseudocode).

\subsection{Techniques}
\label{methods_techniques_sec}
Once a UFH has been generated there are various techniques that can take advantage of it. In this section, we present a simple caching technique and describe two others from the literature.

\subsubsection{Functional Equivalence Cache (FEC)}
\label{methods_fec_technique_sec}

This technique consists of simply caching the fitnesses by the hash, allowing the evolutionary search to skip evaluations. Method \ref{alg:fec} illustrates FEC in the context of regularized evolution.

\begin{algorithm}
\caption{Regularized Evolution \alghlight{with FEC (\alghlightcolor)}}
\label{alg:fec}
\small
\begin{algorithmic}
\STATE $population \gets$ empty queue
\STATE \alghlight{$cache \gets $ empty hash table} \algspacedcomment{Candidate hash to fitness.}
\STATE $n = 0$  \algspacedcomment{Number of sampled candidates.}
\STATE \algcomment{P is the population size (a meta-parameter).}
\WHILE{$|population| < P$}
    \STATE $seed \gets$ RandomCandidate()
    \color{\alghlightcolor}
    \STATE $key \gets $ UnifiedFunctionalHash($seed$)
    \IF{$key$ in $cache$}
        \STATE $seed.fitness \gets cache[key]$  \algspacedcomment{Skips evaluation.}
    \ELSE
    \color{black}
        \STATE $seed.fitness \gets$ Evaluate($seed$)
    \color{\alghlightcolor}
        \STATE $cache[key] \gets seed.fitness$
    \ENDIF
    \color{black}
    \STATE $population.enqueue(seed)$
    \STATE $n = n + 1$
\ENDWHILE
\STATE \algcomment{N is the total number of candidates (a meta-parameter).}
\WHILE{n < N}
    \STATE \algcomment{T is the tournament size (a meta-parameter).}
    \STATE $tournament \gets$ RandomSubset($population$, size=$T$)
    \STATE $parent \gets $ CandidateWithBestFitness($tournament$)
    \STATE $child \gets$ Mutate($parent$)
    
    \color{\alghlightcolor}
    \STATE $key \gets $ UnifiedFunctionalHash($child$)
    \IF{$key$ in $cache$}
        \STATE $child.fitness \gets cache[key]$  \algspacedcomment{Skips evaluation.}
    \ELSE
    \color{black}
        \STATE $child.fitness \gets$ Evaluate($child$)
    \color{\alghlightcolor}
        \STATE $cache[key] \gets child.fitness$
    \ENDIF
    \color{black}
    \STATE $population.enqueue(child)$
    \STATE $population.dequeue()$ \algspacedcomment{Remove oldest.}
    \STATE $n = n + 1$
\ENDWHILE
\STATE \textbf{Return} CandidateWithBestFitness($population$)
\end{algorithmic}
\end{algorithm}

\subsubsection{Other Techniques}
\label{methods_literature_techniques_sec}

We will also apply UFH to two techniques from the literature, the Functional Change Mutator (FCM) and tabulist.

\textbf{FCM} replaces the single mutation on the child with a retry loop that accumulates mutations on the child until it is different from its parent \cite{singh2006comparison}. With UFH, the child is guaranteed to be \emph{functionally} different; this means it is possible for the retry loop to accumulate silent structural mutations. Method~\ref{alg:fcm} in Supplementary Section~\ref{suppl_fcm_sec} shows the details.

\textbf{Tabulist} is a method of steering the search process away from over-explored areas. A table keeps track of the number of times recent candidates have been seen. If a candidate has been seen more than a given number of times, it is considered over-explored. Each new child is mutated until it is no longer in an over-explored area. Mutations are cumulative, as in FCM. Method~\ref{alg:tabulist} in Supplementary Section~\ref{suppl_tabulist_sec} shows the details.

\section{Experimental Setups}
\label{setups_sec}

We will test the effectiveness of the unified functional hashing method (UFH) by applying the techniques in Section~\ref{methods_sec} on four experimental setups from the literature, namely: MNAS \cite{tan2019mnasnet,tan2019efficientnet}, \mbox{AutoML-Zero} \cite{real2020automl}, \mbox{AutoRL} \cite{co2021evolving}, and \mbox{NAS-Bench-101} \cite{ying2019bench}. These choices were motivated in Section~\ref{intro_sec}. In this section, we will describe the tasks in question, corresponding search spaces, and additional details required to reproduce our experiments. The details will include any minor UFH adaptations that were needed for each experimental setup. Generally, we did not modify the experimental setups from the original baselines unless it was necessary, in which case we describe the difference.


\subsection{Setup A: MNAS (Neural Architecture Search)}
\label{mnas_setup_sec}

\begin{figure}[ht]
\centering
\includegraphics[width=\linewidth]{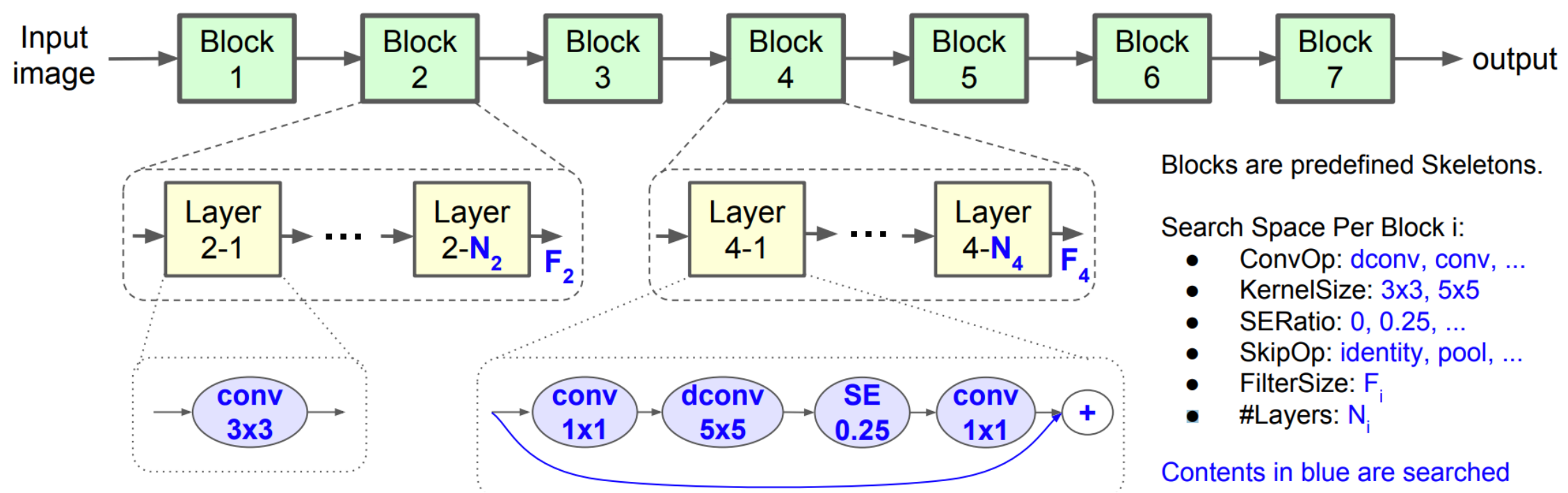}
\caption{Experimental Setup A. Neural networks in the MNAS search space are structured hierarchically into blocks (green), layers (yellow), and ops (blue). The search algorithm must optimize the elements in blue and the number of layers in yellow. Figure from \citeinline{tan2019mnasnet}.}.
\label{mnas_setup_fig}
\end{figure}

MNAS searches for neural architectures for image classification. It aims to optimize the accuracy and latency of models. These two objectives are combined into a single fitness quantity to be maximized, which represents a concrete trade-off between the two objectives. The search space consists of neural networks that stack a fixed number of \emph{blocks}, each block consisting of a \emph{layer} repeated $K$ times. Each layer contains neural network operations such as convolutions and skip connections. The search method must optimize the value $K$, the layers within a cell, and their connectivity, under the constraint that all cells in a block are identical, as indicated in Figure~\ref{mnas_setup_fig} \cite{tan2019mnasnet}.

\experimentdetails[Details]{For the hashable outputs, we used the softmax logits of the resulting neural network, rounded to the fourth decimal place (without tuning). The MNAS search space was kept identical to the original paper. The MNAS search space was kept identical to the original paper. For the search algorithm, the original paper uses reinforcement learning, which is out of our scope. We instead used regularized evolution with population size of $P=100$ and tournament size of $T=10$ without hyperparameter optimization. For the mutations, we randomly alter a single parameter in a block of the model by uniformly randomly sampling a new value (see original paper for architectural details).}

\subsection{Setup B: AutoML-Zero}

AutoML-Zero searches for small ML algorithms, starting from scratch and only using basic components such as addition and multiplication. The candidate algorithms are represented directly as code. Each candidate consists of three functions: \textcode{InitializePass}, \textcode{ForwardPass}, and \textcode{BackwardPass}\footnote{\citeinline{real2020automl} uses instead the terms \textcode{Setup}, \textcode{Predict}, and \textcode{Learn}, respectively.}. Each function consists of a sequence of assembly-like imperative instructions that act on a small virtual memory with scalar, vector, and matrix types (an instruction may be, for example, ``add scalar addresses 0 and 5 and write the output to scalar address 2''). The fitness of a candidate is its validation accuracy after training on a small classification dataset. The search method must discover the sequence of instructions for all three functions, including their operations and input and output addresses, starting from empty code \cite{real2020automl}. Figure~\ref{figure:automl-zero-setup} shows a program discovered from an actual search experiment.

\begin{figure}[h]
\begin{code}{0.48\textwidth}{2.75in}{codebackground}
\codeline{\codecomment{sX/vX/mX = scalar/vector/matrix at address X.}}
\codeline{\codecomment{``gaussian'' produces Gaussian IID random numbers.}}
\codeskip
\codeline{\codedef{def} InitializePass():}
  \codeline{\codetab \codecomment{Initialize variables.}}
  \codeline{\codetab m1 = gaussian(-1e-10, 9e-09) \codecomment{1st layer weights}}
  \codeline{\codetab s3 = 4.1 \codecomment{Set learning rate}}
  \codeline{\codetab v4 = gaussian(-0.033, 0.01) \codecomment{2nd layer weights}}    \codeskip
\codeline{\codedef{def} ForwardPass(): \codecomment{v0=features}}
  \codeline{\codetab v6 = dot(m1, v0) \codecomment{Apply 1st layer weights}}
  \codeline{\codetab v7 = maximum(0, v6) \codecomment{Apply ReLU}}
  \codeline{\codetab s1 = dot(v7, v4) \codecomment{Compute prediction}}
\codeskip
\codeline{\codedef{def} BackwardPass(): \codecomment{s0=label}}
  \codeline{\codetab v3 = heaviside(v6, 1.0) \codecomment{ReLU gradient}}
  \codeline{\codetab s1 = s0 - s1 \codecomment{Compute error}}
  \codeline{\codetab s2 = s1 * s3 \codecomment{Scale by learning rate}}
  \codeline{\codetab v2 = s2 * v3 \codecomment{Approx.\ 2nd layer weight delta}}
  \codeline{\codetab v3 = v2 * v4 \codecomment{Gradient w.r.t.\ activations}}
  \codeline{\codetab m0 = outer(v3, v0) \codecomment{1st layer weight delta}} 
  \codeline{\codetab m1 = m1 + m0 \codecomment{Update 1st layer weights}}
  \codeline{\codetab v4 = v2 + v4 \codecomment{Update 2nd layer weights}}
\end{code}
\caption{Experimental Setup B. An example of a point in the AutoML-Zero search space. This evolved code can be seen to implement a neural network with backpropagation by gradient descent. Figure from \citeinline{real2020automl}.}
\label{figure:automl-zero-setup}
\end{figure}


\experimentdetails[Details]{For the hashable outputs, we used the errors from 10 training and 10 validation examples. $m_{bits} = 27$.
The AutoML-Zero search space was kept identical to that of Section 4.2 of the original paper.}

\subsection{Setup C: AutoRL}
\label{autorl_setup_sec}

\begin{figure}[ht]
\begin{subfigure}{.47\textwidth}
    \centering
    \includegraphics[width=\linewidth]{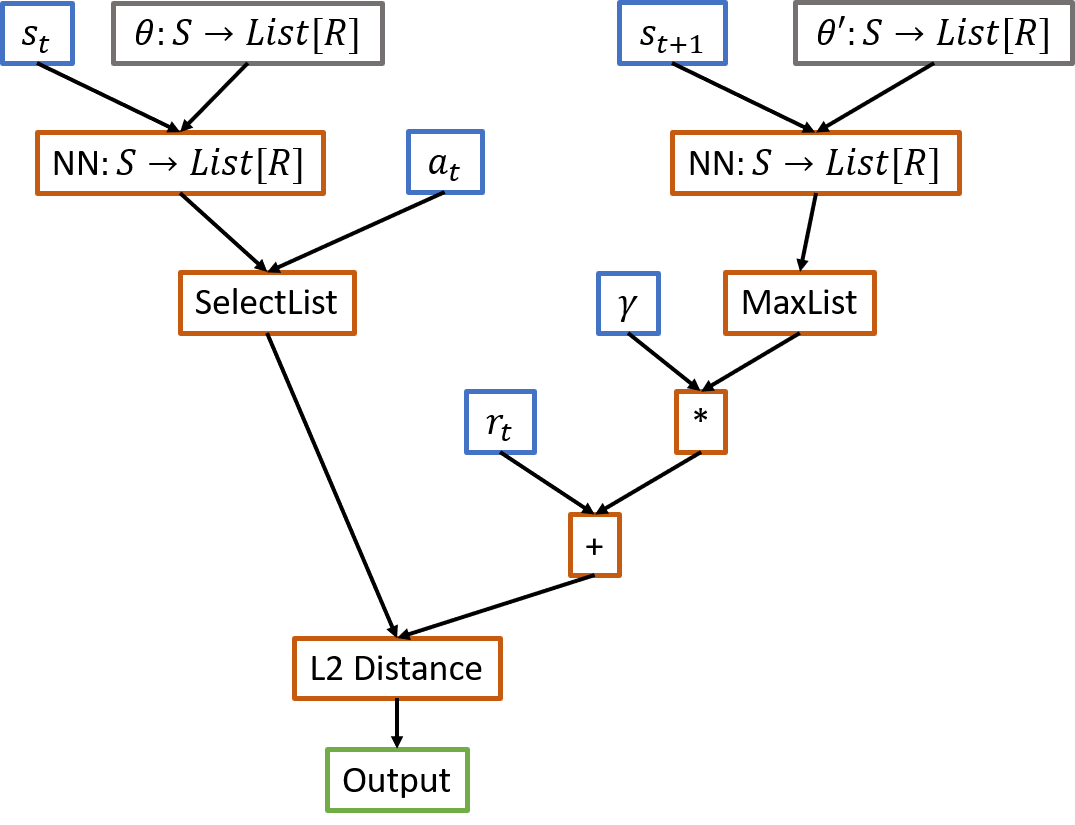}
\end{subfigure}
\caption{Experimental Setup C. Example of a point in the AutoRL search space. This particular compute graph represents the DQN algorithm by encoding the loss function $L = (Q(s_t, a_t) - (r_t + \gamma \times max_a Q_{targ}(s_{t+1}, a)))^2$, where Q is the Q-function, $s$ the state, $a$ the action, and $\gamma$ the discount factor. Figure from \citeinline{co2021evolving}.}
\label{autorl_setup_fig}
\end{figure}

AutoRL searches for reinforcement learning (RL) algorithms \cite{parkerholder2022survey}. We will use as our baseline the specific setup of \citeinline{co2021evolving}, where the goal is to find algorithms that produce high rewards when training RL agents. Namely, a candidate algorithm's fitness is the mean reward obtained when it is used to train an agent on standard RL benchmark environments like the ``cartpole'' \cite{openai2016gym}. The search space consists of possible RL loss functions, each corresponding to an algorithm. These loss functions are represented as directed acyclic compute graphs with an operation at each vertex. Inputs include samples from the replay buffer and neural network parameters.  An example can be seen in Figure~\ref{autorl_setup_fig}. The search algorithm must assign to each vertex an operation (from a predefined list of options) and must determine the connectivity of the vertices in the graph.

\experimentdetails[Details]{For the hashable outputs, we evaluate the loss graph on 4096 uniformly sampled inputs. Mantissas were truncated at $m_{bits}=8$. For the purposes of hashing, we used states and rewards uniformly sampled from [-100, 100] and actions uniformly sampled from the discrete action space (see discussion on ``fake data'' for motivation). The AutoRL search space and search algorithms were kept identical to the original paper. We evaluate loss graphs using three gym environments: CartPole-v0 and LunarLander-v2 from OpenAI Gym suite \cite{openai2016gym}, and DynamicObstacle-6x6 from the MiniGrid suite \cite{gym_minigrid}. We searched over graphs with a maximum of 20 nodes, excluding inputs or parameter nodes. For regularized evolution, we used a population size of 300, tournament size of 25, mutation rate of 0.95. We ran the evolution on 300 CPUs for 7 days to ensure sufficient convergence.}

\subsection{Setup D: NAS-Bench-101}

\begin{figure}[ht]
\centering
\includegraphics[width=\linewidth,trim={0 9cm 4mm 0},clip]{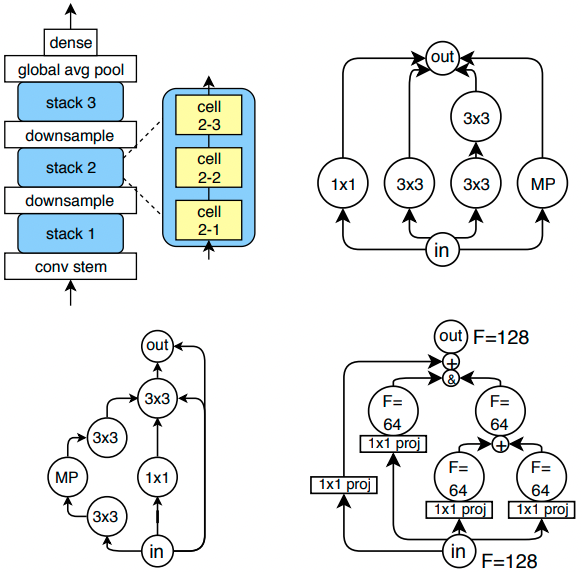}
\caption{Experimental Setup D. NAS-Bench-101 search space. LEFT: Each point in the space encodes an image classifier with a fixed outer skeleton of stacked, identical cells (yellow). The points differ in the inner architecture of the cells. RIGHT: An example of a cell with an Inception-like architecture. Figure from \citeinline{ying2019bench}.}
\label{figure:nasbench_setup_fig}
\end{figure}

NAS-Bench-101, like MNAS, is also a neural architecture search task for image classification. The fitness objective is the classification accuracy. The search space consists of compute graphs of neural network operations, like convolutions or pooling. This graph forms a so-called \emph{cell} that gets stacked a fixed number of times to build an image classifier. Figure~\ref{figure:nasbench_setup_fig} provides a visual description. Unlike MNAS, NAS-Bench-101 has a tiny search space (only about 400k unique architectures), all of which have been previously evaluated and tabulated by the creators of the benchmark. As a result, searching this space is extremely fast as no training needs to be done within the AutoML loop. Training times are included in the benchmark's table, so they can be properly accounted for during simulations. The task for the search method is to discover the operations at each vertex of the compute graph, as well as the graph connectivity \cite{ying2019bench}.


\experimentdetails[Details]{For the hashable outputs, we would ideally use the logits, but this is impossible in NAS-Bench-101 as we are not training neural networks but rather querying the results of pre-computed training and validation runs. Fortunately, in addition to the full training results (108 epochs, whose validation accuracy we use as fitness), the benchmark's table includes accuracy values for shorter training runs. We use 4 accuracy values for the shortest run (4 epochs) as hashable outputs. Mantissas were truncated at $m_{bits}=24$. We assumed hashing to cost 10s of the budget, which is likely to be an overestimate and therefore bias the results against our method. For regularized evolution and classic tournament selection we performed 500 trials of each algorithm, whereas for reinforcement learning with PPO we performed 1000 trails. The search space, the application of regularized evolution, and the application of tournament selection are all identical to those in \citeinline{ying2019bench}, including the meta-parameters and simulated experiment durations. For PPO, we did not have the original parameters, so we used the default configuration in \citeinline{peng2021pyglove}.
}

\section{Results}
\label{results_sec}

\begin{figure*}[ht]
\begin{subfigure}{.33\textwidth}
    \centering
    \includegraphics[width=0.9\linewidth]{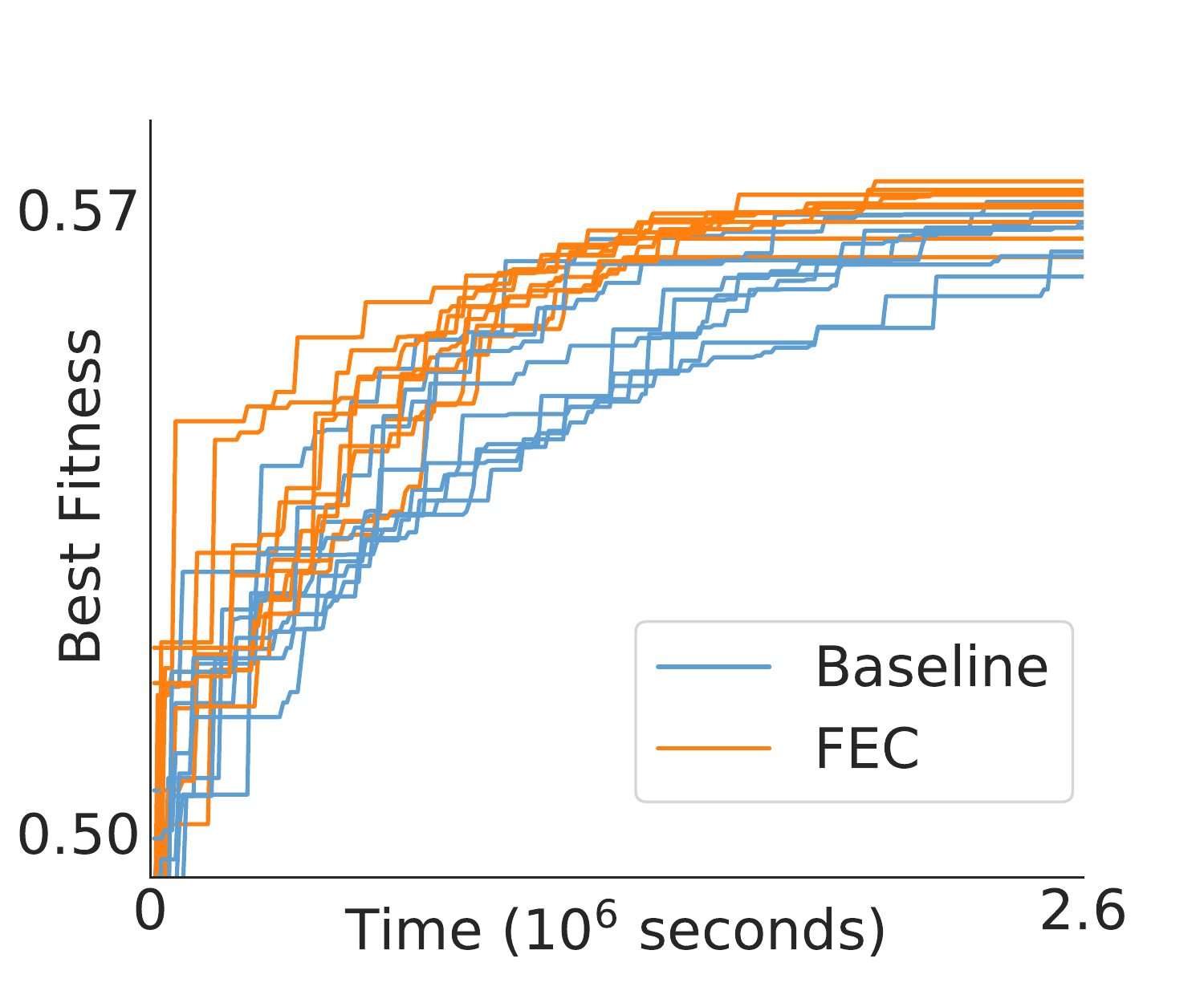}
\end{subfigure}
\begin{subfigure}{.33\textwidth}
    \centering
    \includegraphics[width=0.9\linewidth]{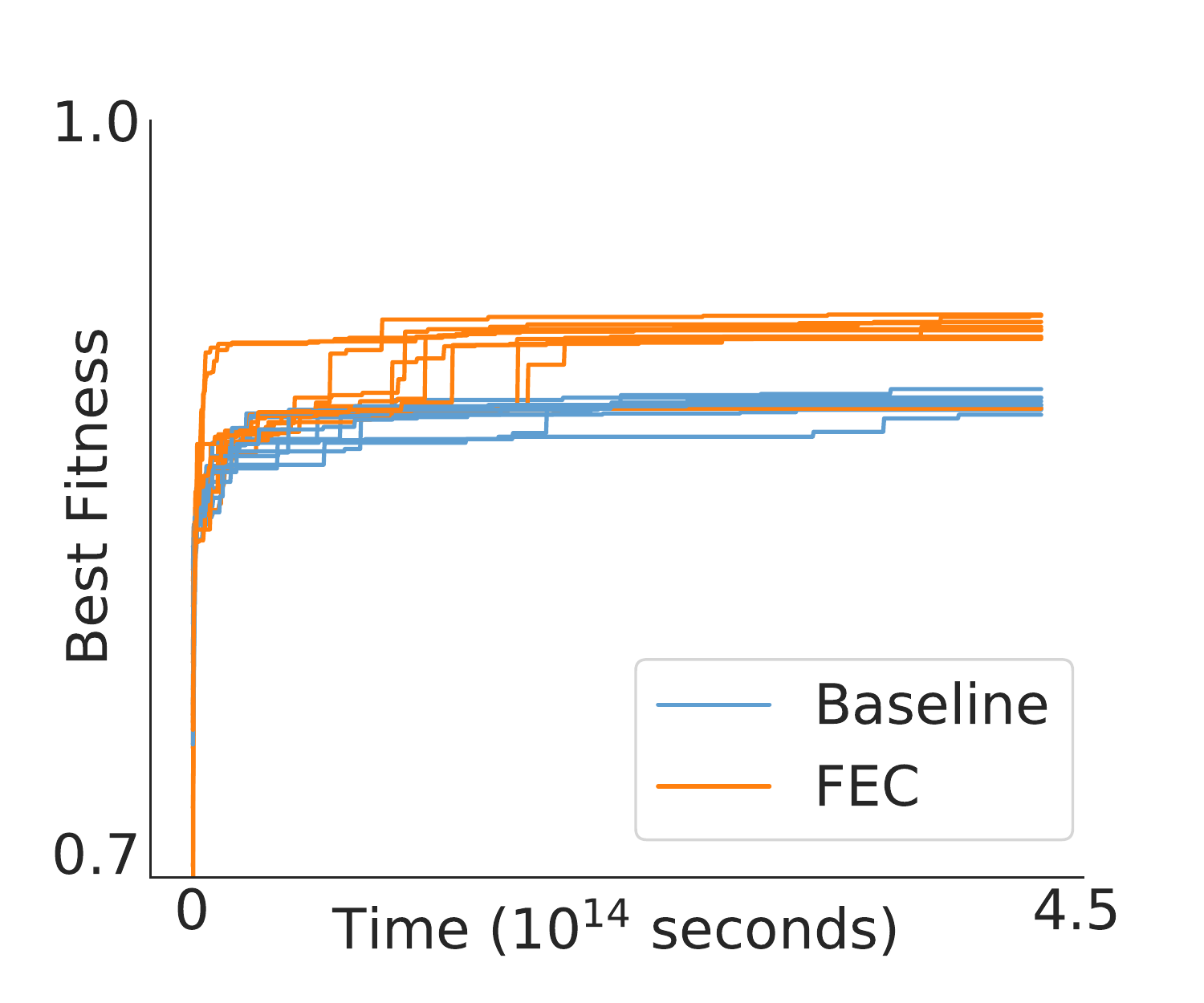}
\end{subfigure}
\begin{subfigure}{.33\textwidth}
    \centering
    \includegraphics[width=0.9\linewidth]{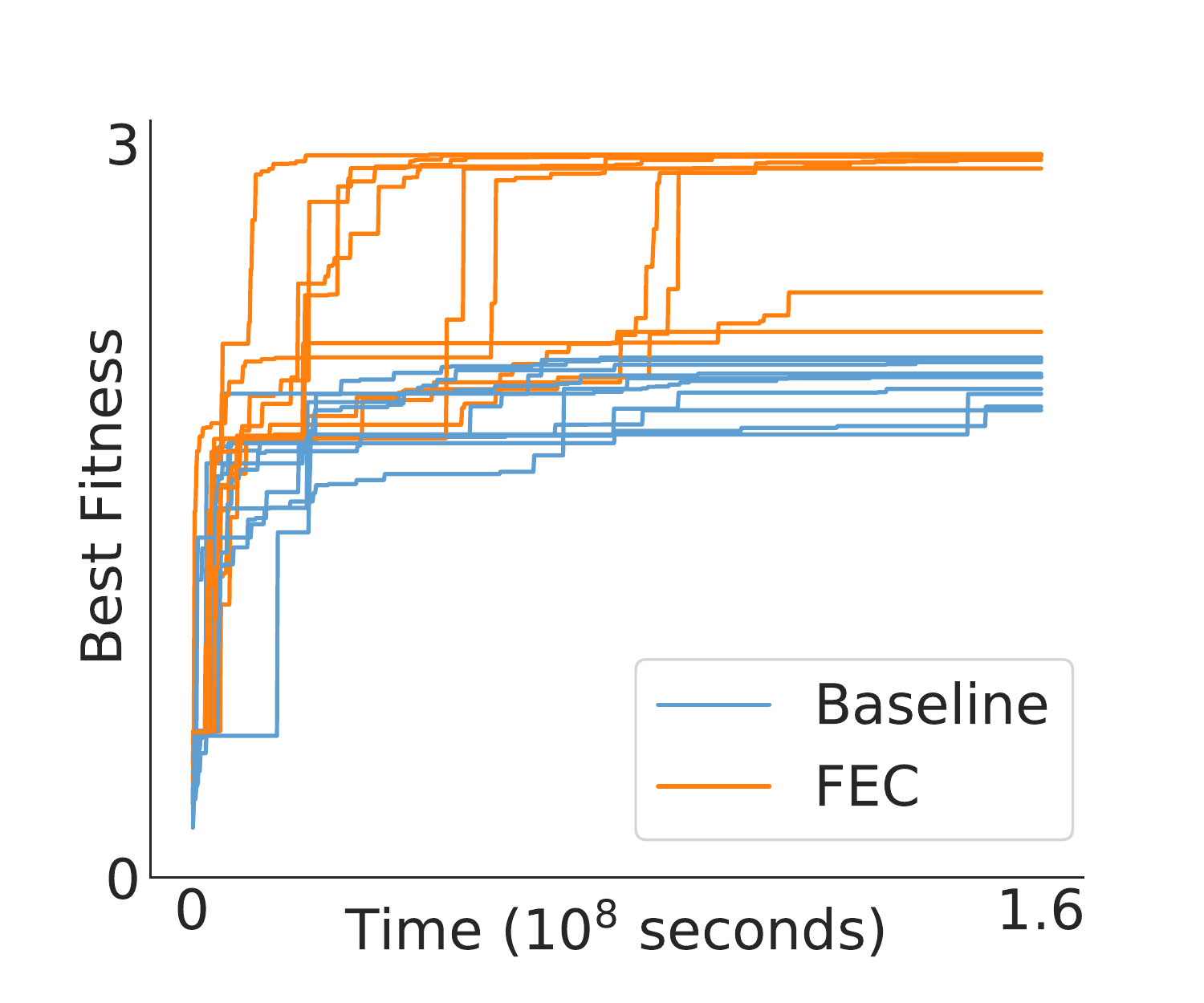}
\end{subfigure}

\begin{subfigure}{.33\textwidth}
    \centering
    \includegraphics[width=0.9\linewidth]{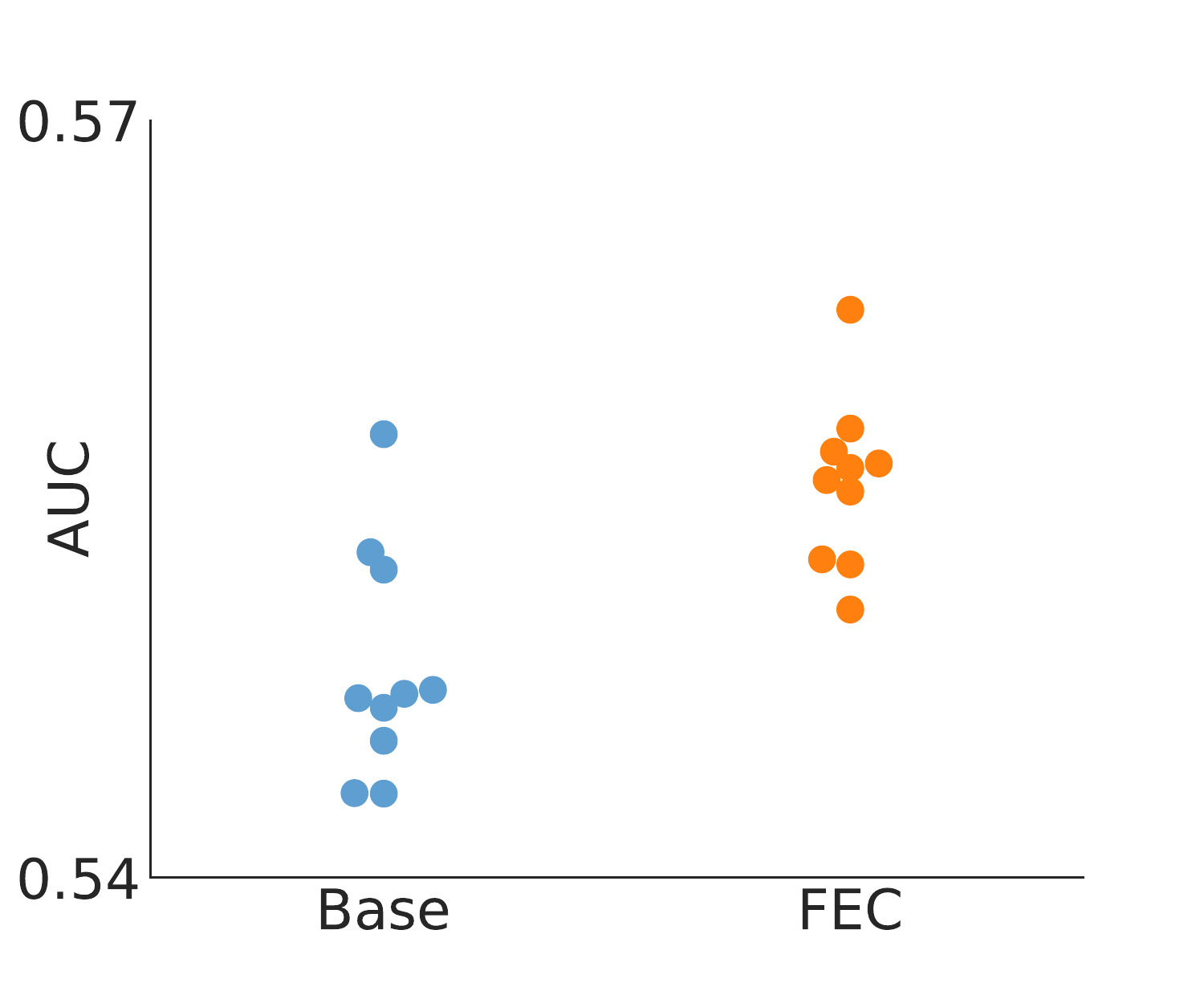}
\end{subfigure}
\begin{subfigure}{.33\textwidth}
    \centering
    \includegraphics[width=0.9\linewidth]{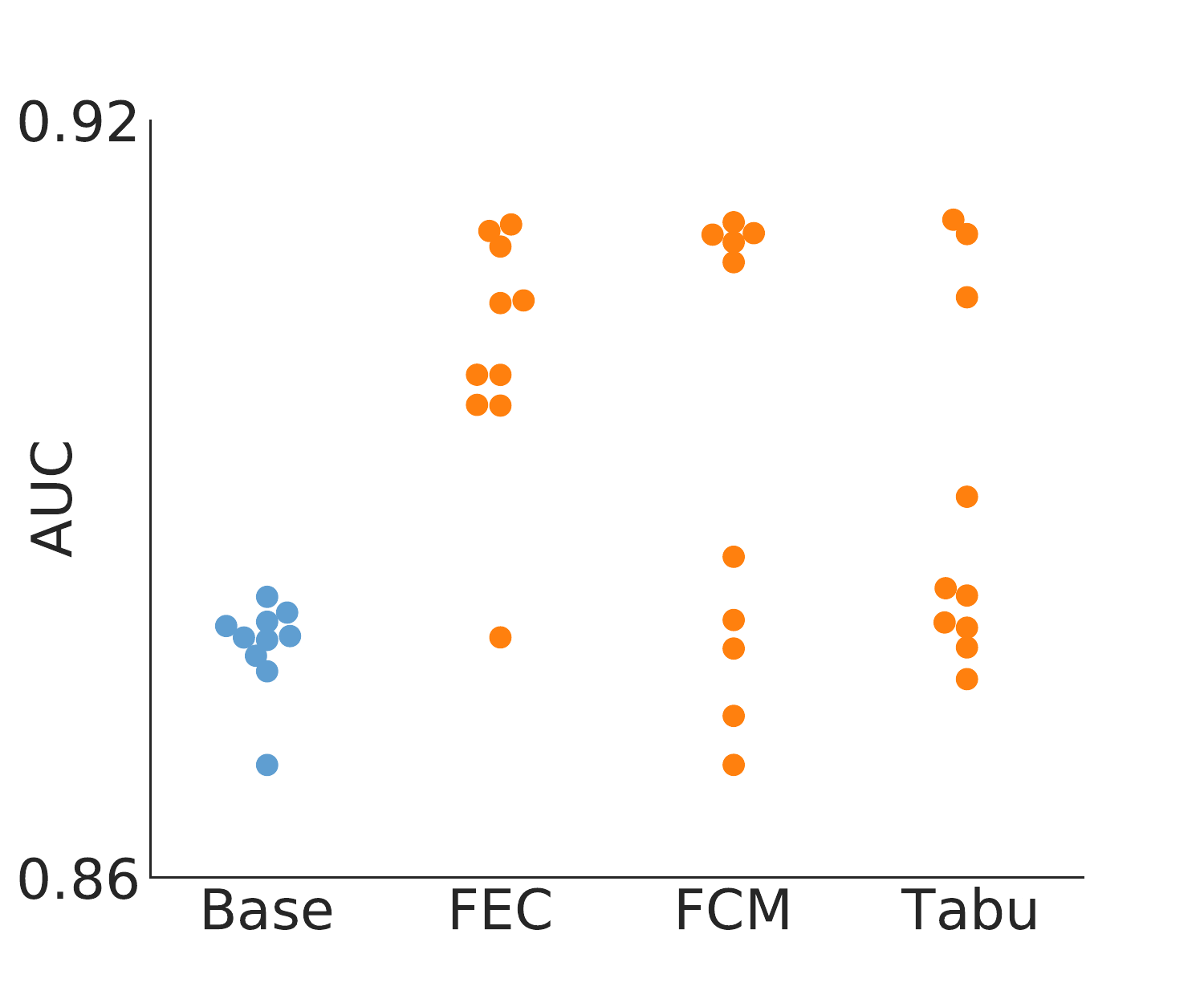}
\end{subfigure}
\begin{subfigure}{.33\textwidth}
    \centering
    \includegraphics[width=0.9\linewidth]{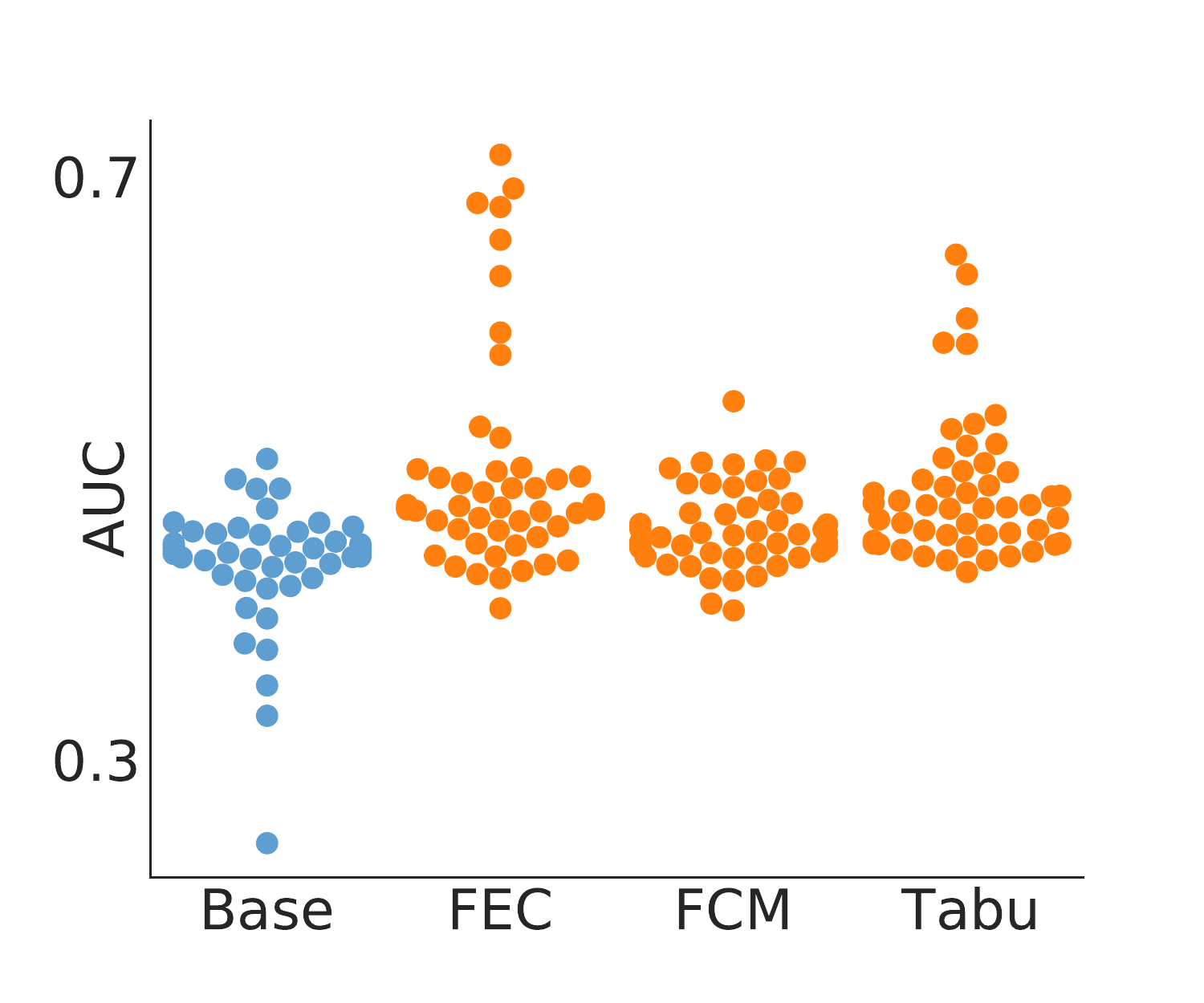}
\end{subfigure}
\caption{Improvements due to UFH-based techniques. From left to right, the columns correspond to MNAS, AutoML-Zero, and AutoRL. TOP ROW: FEC results; each line represents the full time-course of the fitness of a search  experiment (higher is better). BOTTOM ROW: comparison of all techniques; each point marks the AUC of the fitness curve of an experiment (higher is better).}
\label{metatrain_fig}
\end{figure*}

Our unified functional hash (UFH), when applied through the functional equivalence cache (FEC) technique resulted in faster search across all experimental setups, with equal or better final results. To demonstrate this, we performed A/B tests controlling for compute budget. The top row of Figure~\ref{metatrain_fig} shows the results of the time-course of the experiments. FEC curves tend to dominate non-FEC curves. Thus, FEC allows discovering algorithms faster. In all setups, FEC-discovered algorithms were at least as good as baseline ones. In some setups they were strictly superior (that is, once the experiment was complete, the best discovered algorithm of an experiment with FEC was better than the best discovered algorithm without FEC). We highlight that the experiment termination criterion was not chosen by us but by the original baseline studies.

When applying UFH through other techniques, we obtained outcomes that were generally better and never worse. The bottom row of Figure~\ref{metatrain_fig} summarizes many experiments across all three techniques: FEC, functional change mutation (FCM), and tabulist. In the MNAS setup, we only ran FEC experiments at scale, due to the large resources required for each single experiment. In the figure, each experiment is summarized with a single number, the experiment's area-under-curve (AUC). We define the AUC to be the integral of the fitness curve. That is, $AUC = 1/T \int_{0}^{T} f_{(t)} \,dt $, where T is the duration of the experiment. As the figure shows, the AUC from applying each of the three techniques tends to be higher, indicating the methods produce better results and/or discover them faster. Full time-courses for all techniques can be found in the Supplementary Materials.

\begin{figure*}[ht]
\begin{subfigure}{.33\textwidth}
    \centering
    \includegraphics[width=0.9\linewidth]{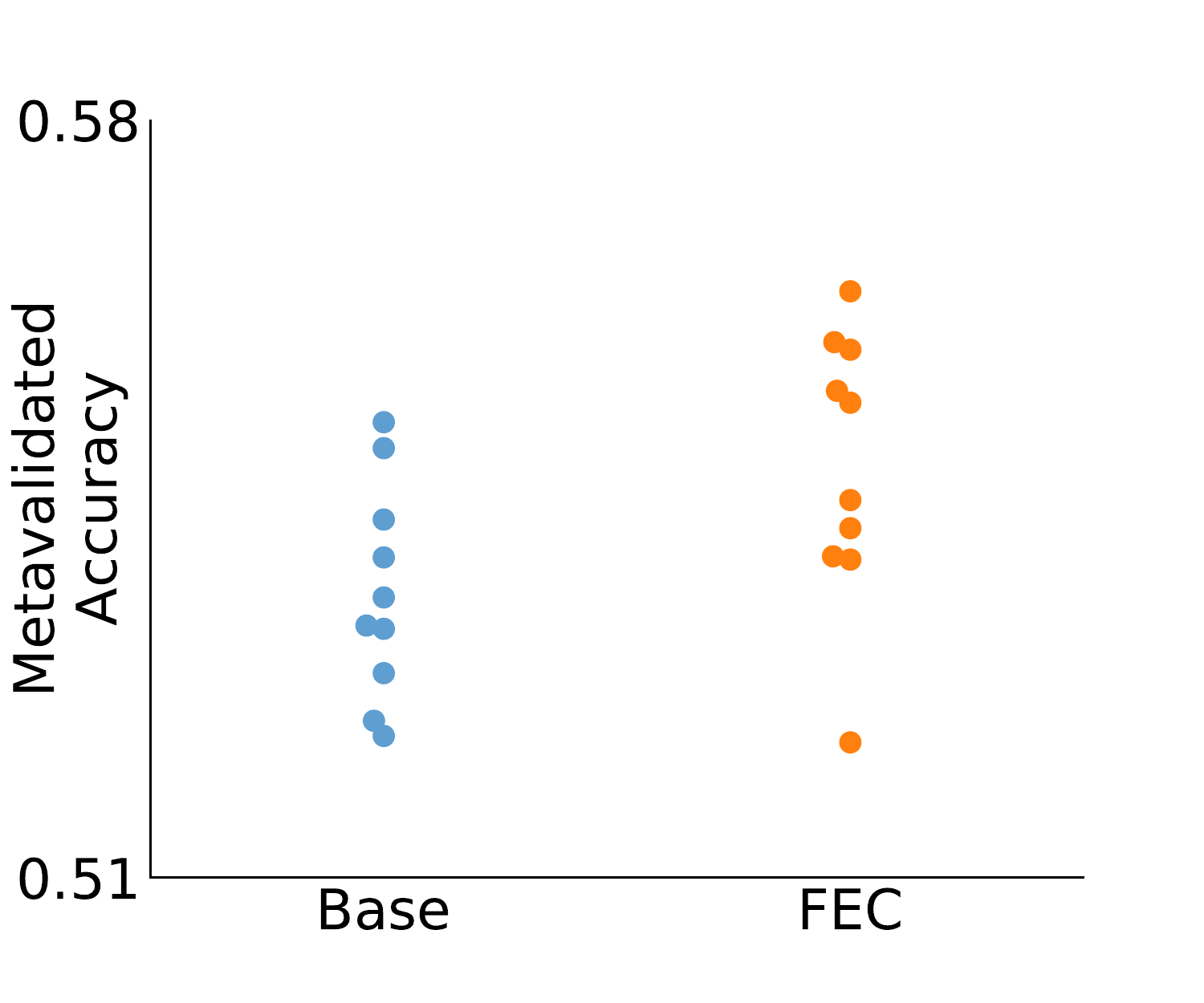}
    \caption{}
    \label{metavalid_mnas_subfig}
\end{subfigure}
\begin{subfigure}{.33\textwidth}
    \centering
    \includegraphics[width=0.9\linewidth]{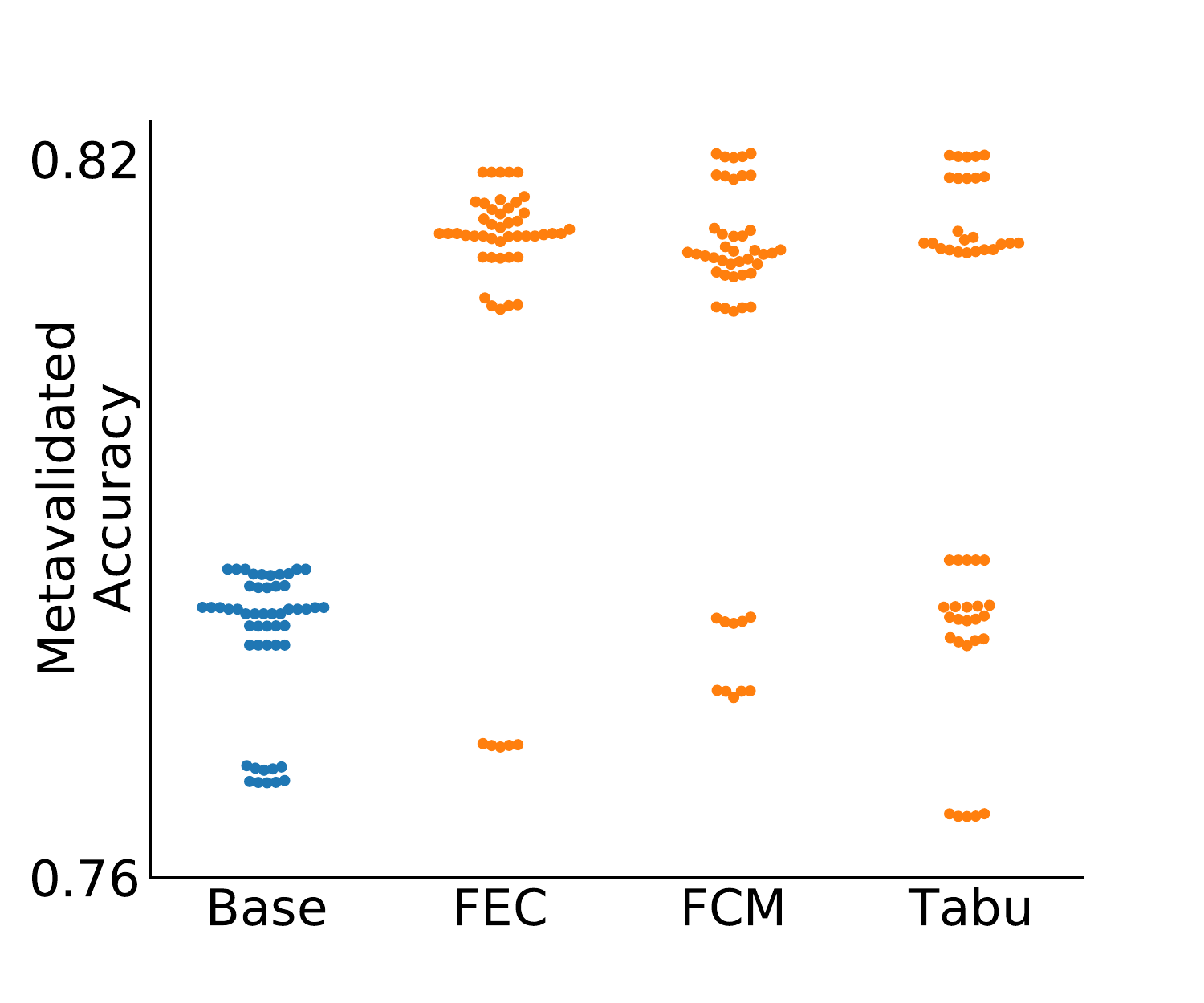}
    \caption{}
    \label{metavalid_amlz_subfig}
\end{subfigure}
\begin{subfigure}{.33\textwidth}
    \centering
    \includegraphics[width=0.9\linewidth]{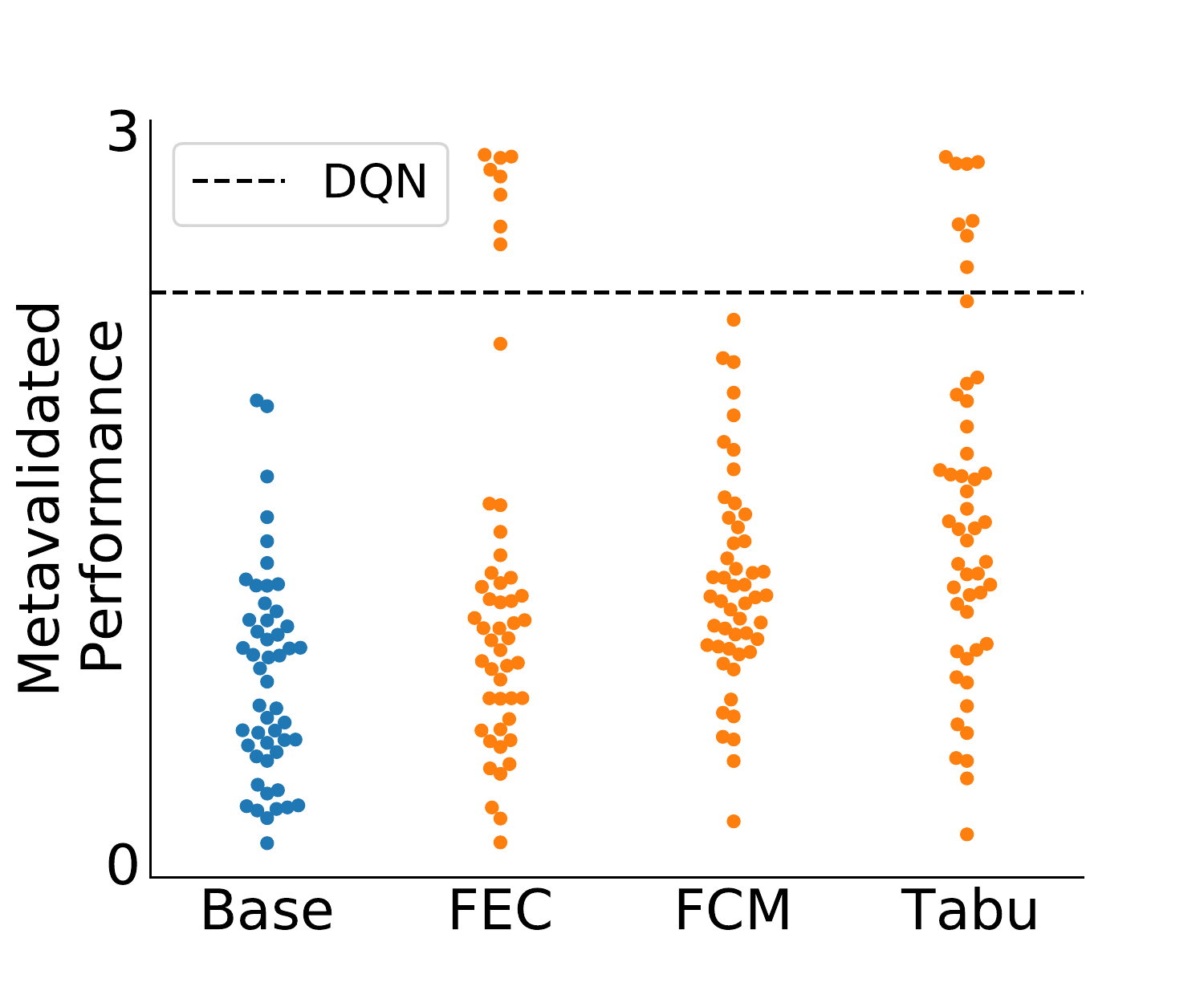}
    \caption{}
    \label{metavalid_autorl_subfig}
\end{subfigure}
\caption{Reevaluation of discovered algorithms on unseen data. LEFT: MNAS, MIDDLE: AutoML-Zero, RIGHT: AutoRL. Each point marks the final accuracy of the best model discovered in the final population, evaluated on test data. Best models were chosen without resorting to the test set. For MNAS, we sampled the population in the middle of the experiment instead of the final one, to highlight the differences in search speed. For AutoRL, the performance of the hand-crafted DQN algorithm is shown for reference. In all cases, results support the qualitative differences seen in Figure~\ref{metatrain_fig}.}
\label{metavalid_fig}
\end{figure*}

Re-evaluation of discovered algorithms on unseen data confirms the results above. This is an important verification because it shows that the improvement was not just due to overfitting. The search process can overfit algorithms to the data just like training an ML model can overfit parameters. In ML, this is due to the model having too much capacity and can be resolved by either directly reducing the complexity of the model or adding regularization to the training. Similarly, in AutoML, overfitting can be resolved by reducing the search space or by regularizing the search process. Reducing the search space is outside our scope, as we assume that the spaces have been well crafted in their original papers. However, since FEC/FCM/tabulist directly affect the search process, we need to ensure that they have not resulted in \emph{additional} overfitting. Figure~\ref{metavalid_fig} shows that overfitting is not a concern, as the relationship to the baseline is preserved when reevaluating on unseen data. In all cases, we picked the best algorithms at the end of the experiments to reevaluate. In cases where both methods seem to converge to the same asymptote, we also reevaluated mid-experiment results. We conclude that UFH methods tend to result in faster and/or better discoveries.

\section{Analysis and Discussion}
\label{discussion_sec}

In this paper, we have introduced a method to perform a fast unified functional hash (UFH) of arbitrary learning algorithms, regardless of their representation. We have also demonstrated its applicability to evolutionary AutoML through the functional equivalence caching (FEC) technique. Additionally, we have demonstrated broader potential through two classic techniques: the functional change mutator (FCM) and the tabulist. Ultimately, their effectiveness depends on the search space, but our results give positive evidence of their scope.

In this section, we will consider aspects of hashing that can have a large impact on its application, focusing on the FEC technique. The reason for this focus is that FEC is simple and gave the most widely applicable results. To illustrate our points, we will use NAS-Bench-101, as this setup is the only one that provides the fast experimentation needed here. We will start out by testing the robustness of the FEC method. Then, we will outline the guarantees that FEC provides under certain conditions, followed by the effects of breaking those conditions (distributed search, hash collisions, evaluation noise). In so doing, we will also discuss possible directions for future work. We will also include below guidelines for practitioners wishing to use functional hashing methods.

\subsection{FEC Robustness}
\label{discussion_robustness_sec}

\begin{figure}[ht]
    \centering
    \includegraphics[width=0.9\linewidth]{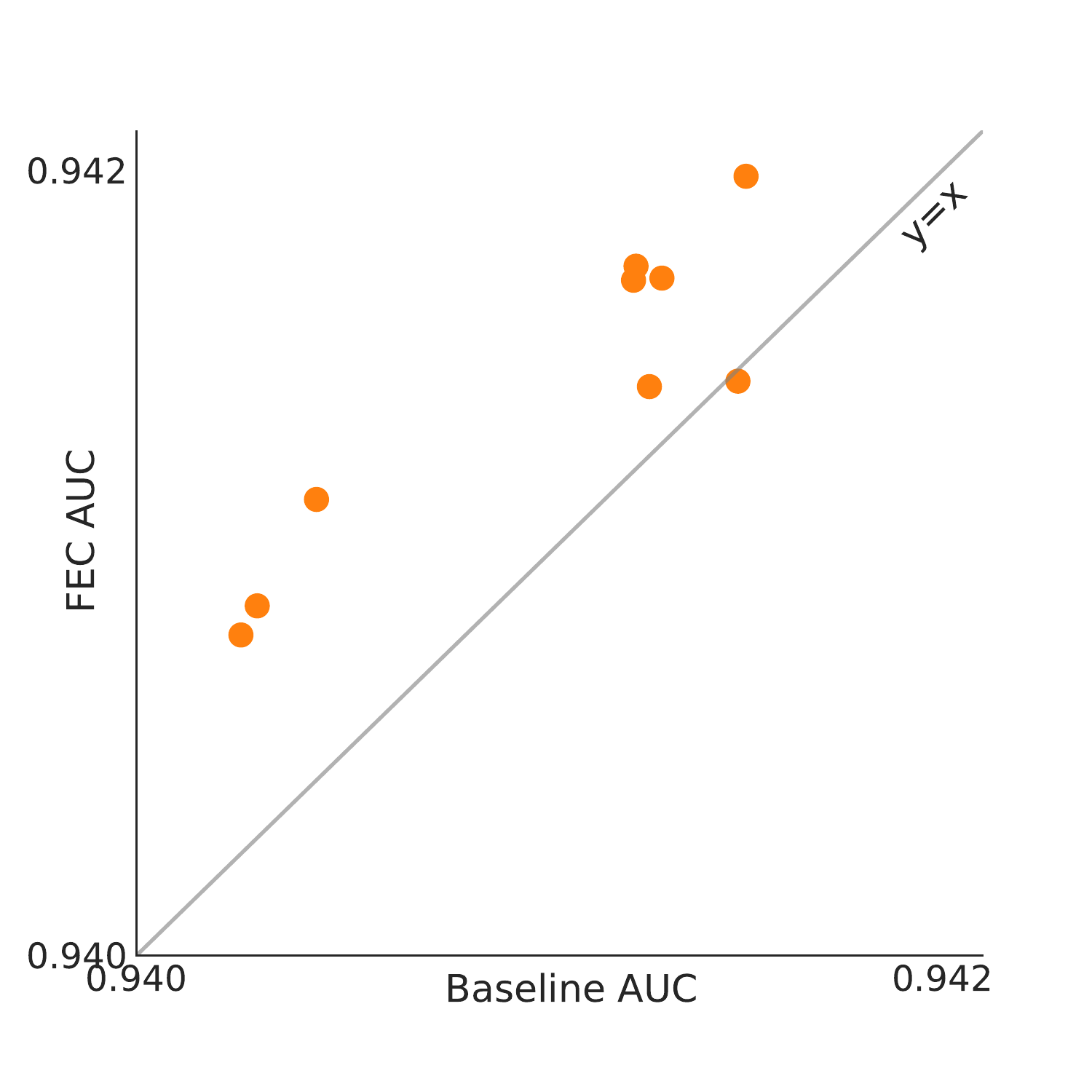}
\caption{Robustness to the evolutionary hyper-parameters. Each point represents a different hyperparameter choice, the x-coordinate measures the mean AUC of a set of 100 baseline search experiments, and the y-coordinate measures the mean AUC of 100 experiments that use FEC. Most points are above $y=x$, showing that using FEC is preferable across hyperparameters. The set of points is from the Cartesian product of two hyperparameters taking values in the sets, population size = \{25, 50, 100\} and tournament size = \{2, 5, 10\}.}
\label{hyperparameter_robustness_fig}
\end{figure}

FEC does not seem to complicate the application of evolutionary methods. Its basic form does not add any hyperparameters that need to be tuned. Note that the cache size parameter can be chosen to be as large as there is RAM to spare; it does not need tuning with a grid search over many experiments. This is important because AutoML experiments can be slow. For the same reason, it is important that the introduction of FEC does not make the search process more sensitive to the choices of the preexisting evolutionary hyperparameters, namely population size and tournament size.
While NAS-Bench-101 is fast enough for the experiments needed here, its small dataset size and low problem difficulty make distinctions between different search methods harder than setups used in Section~\ref{results_sec}. Nevertheless, FCM and tabulist showed no ill effects and FEC provided an improvement. Figure~\ref{hyperparameter_robustness_fig} shows that this improvement holds for a wide range of search hyperparameters.

\begin{figure*}[ht]
\begin{subfigure}{.35\textwidth}
    \includegraphics[width=1.0\linewidth]{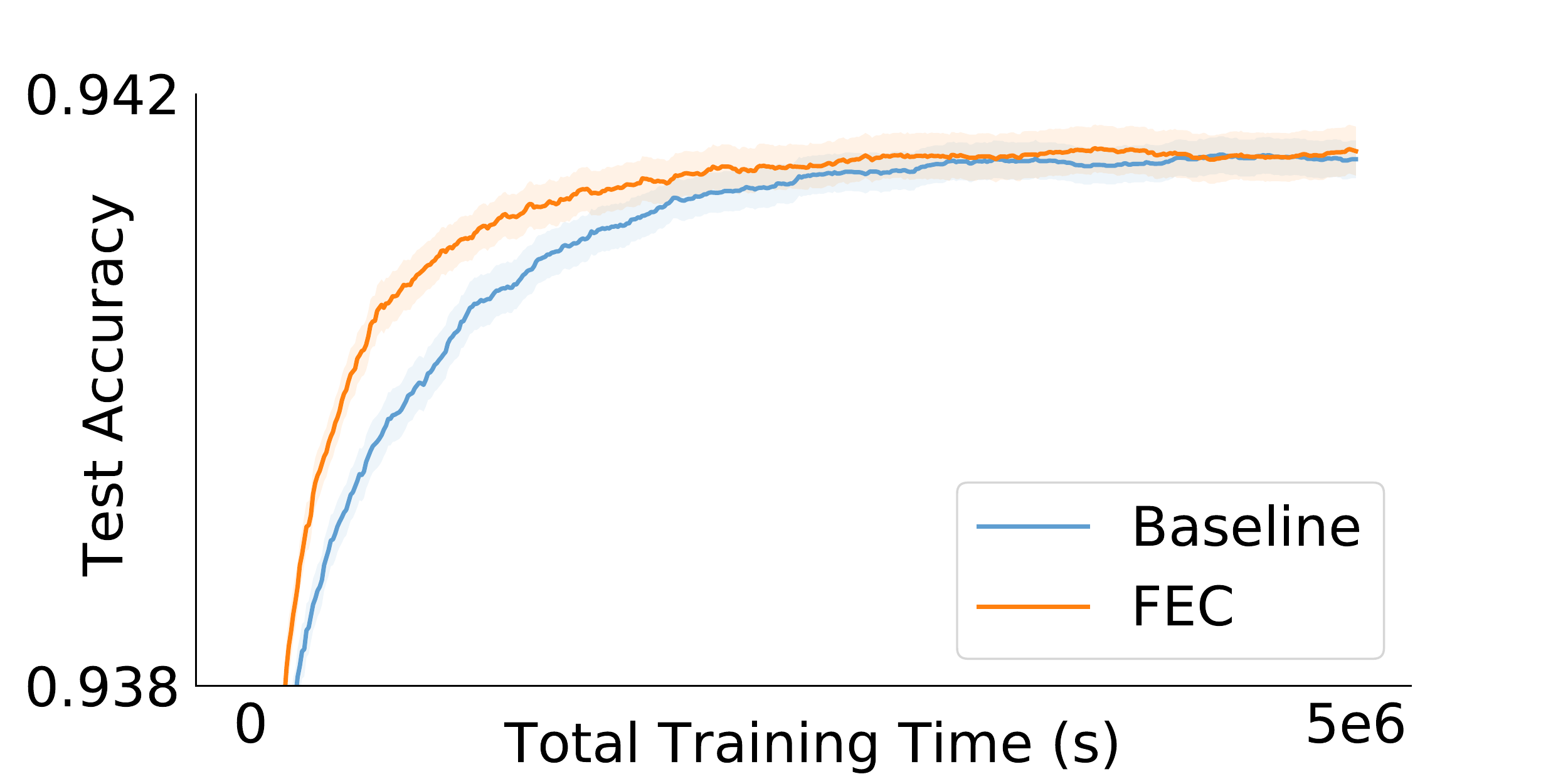}
\end{subfigure}
\begin{subfigure}{.35\textwidth}
    \hspace*{-.6cm}
    \includegraphics[width=1.0\linewidth]{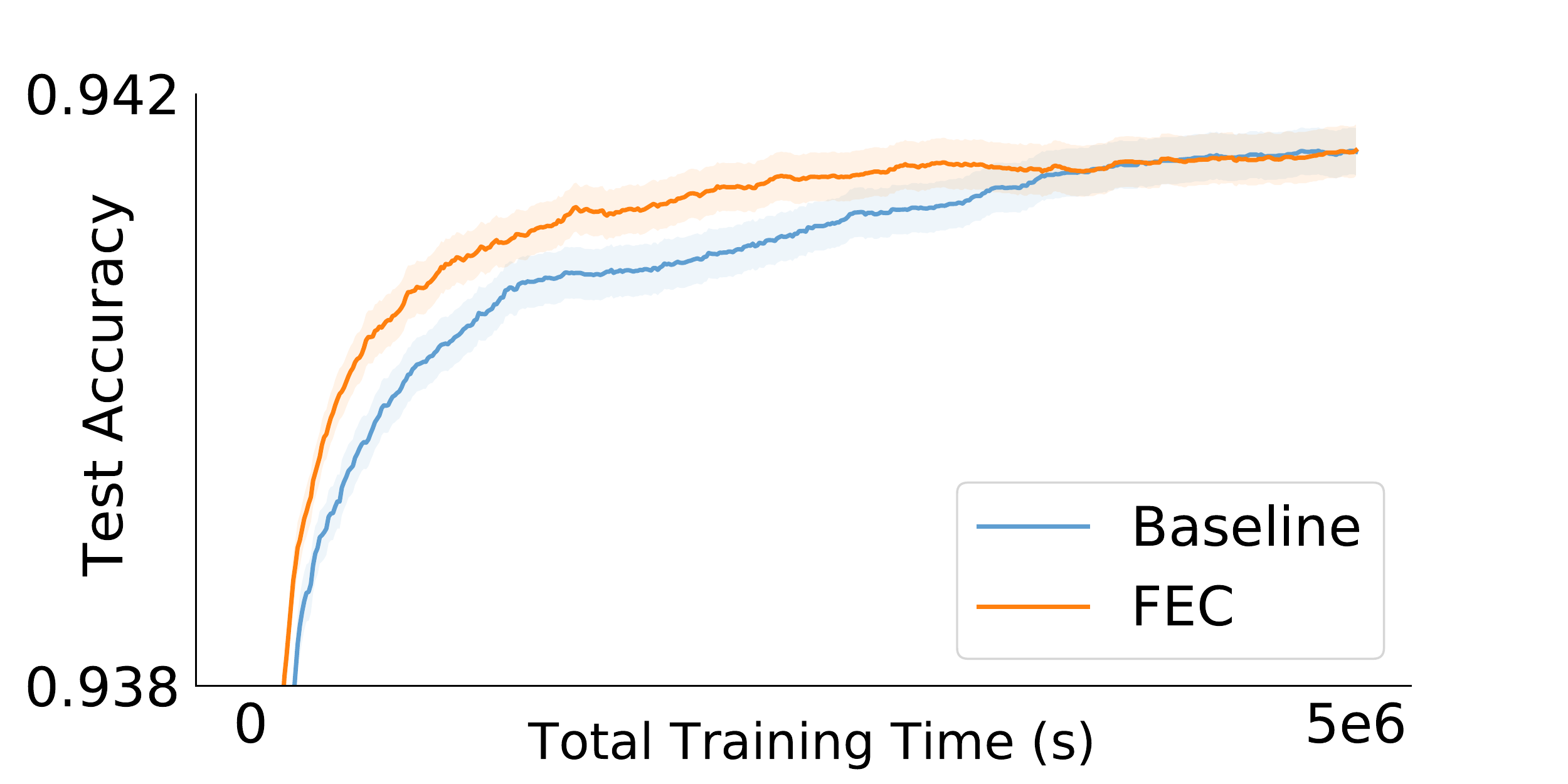}
\end{subfigure}
\begin{subfigure}{.35\textwidth}
    \hspace*{-.9cm}
    \includegraphics[width=1.0\linewidth]{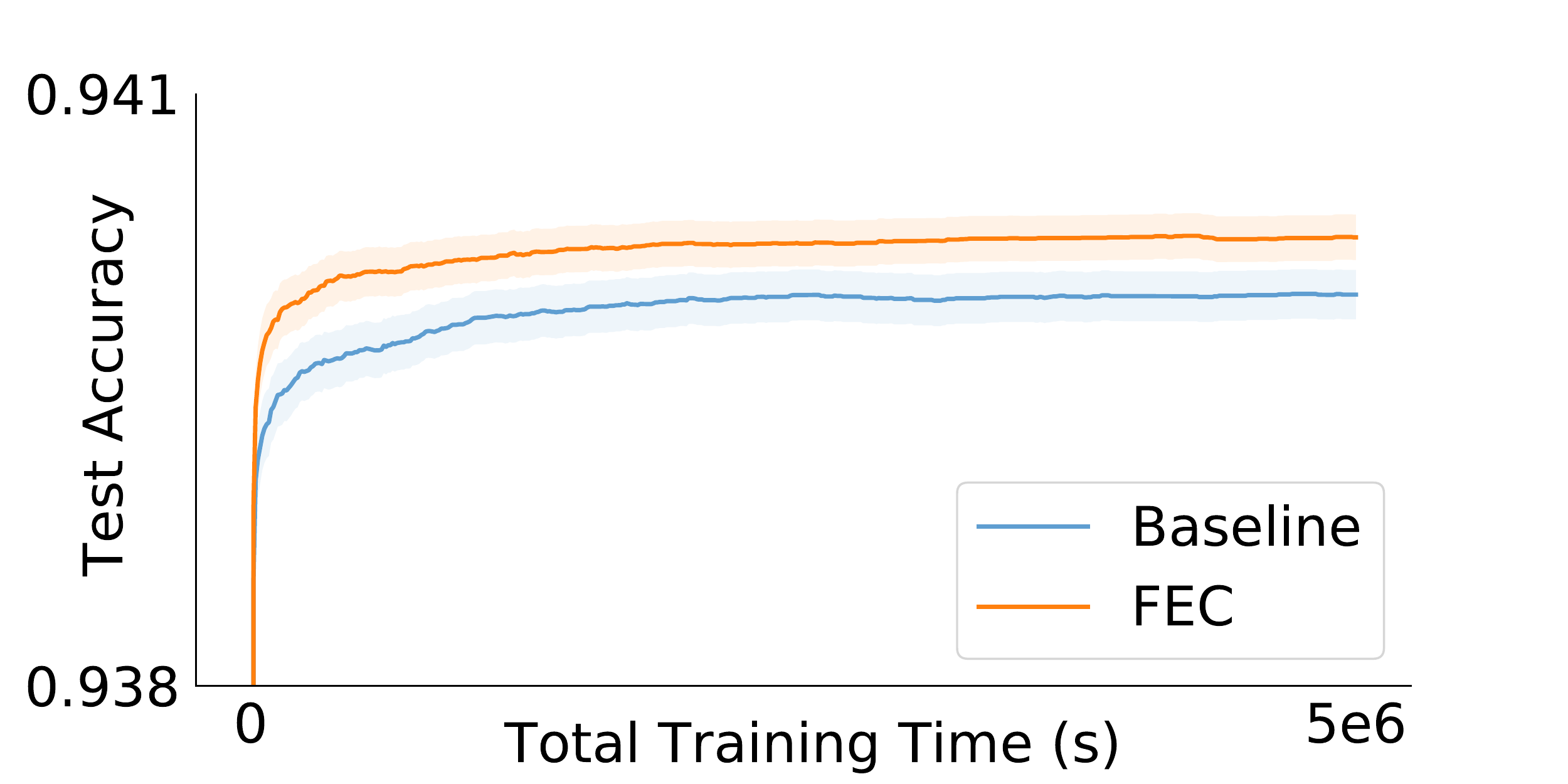}
\end{subfigure}
\caption{Robustness to the search method in the NAS-Bench-101 setup. Lines are the mean across 500 experiments with the shading being $\pm$ 2 SEM. LEFT: Regularized evolution. MIDDLE: Classic tournament selection. RIGHT: reinforcement learning with PPO, as implemented in \citeinline{peng2021pyglove}.}
\label{method_robustness_fig}
\end{figure*}

Moreover, FEC can be applied beyond the regularized evolution search method used thus far, even in very different search methods such as reinforcement learning (RL). For example, Figure~\ref{method_robustness_fig} compares the application of FEC to regularized evolution (LEFT) and to classic tournament selection (MIDDLE). The choice of classic tournament selection was arbitrary but informed by the major classification of evolutionary algorithms into $(\mu, \lambda)$ vs $(\mu+\lambda)$. Loosely speaking, the former type ensures a finite lifetime for all individuals, as does regularized evolution. Classic tournament selection belongs to the latter type, where elitism permits good individuals to live forever. Furthermore, benefits can also be seen if instead of an evolutionary method, we perform the search with reinforcement learning (RL). We observed this by comparing a standard PPO search with one that was augmented with FEC (Figure~\ref{method_robustness_fig}--RIGHT). We used PPO as the chosen RL method because it has been popular in recent AutoML research \cite{zoph2016neural,ramachandran2017swish,tan2019mnasnet}.

\subsection{FEC Guarantees}
\label{fec_guarantee_sec}

FEC is a very direct application of hashing. In addition to its simplicity, it is appealing because it provides faster search speed while interfering minimally with the evolutionary dynamics. Indeed, it guarantees unaltered evolutionary dynamics under the conditions that (i) evaluations are done serially, (ii) there are no cache collisions, and (iii) evaluations are deterministic. That is, given these conditions, for the same random seed, an experiment produces the same sequence of candidates regardless of whether FEC is used. In the remainder of this section, we will consider the effects of breaking each of these conditions.

\subsubsection{Distributed Search} 

In practice, FEC proved advantageous in spite of condition (i) being rarely satisfied in practice. Evaluations in typical AutoML experiments are not serial, as they are distributed by multi-core processors or multi-machine clusters. This was the case for all experiments in Section~\ref{results_sec}. When multiple candidates are being evaluated in parallel, the presence of FEC can alter the order in which they are added into the population because cache hits will finish much faster than the cache misses. (Note that the same is not true if evaluations are serial.)

\begin{figure}[ht]
    \centering
    \includegraphics[width=0.9\linewidth]{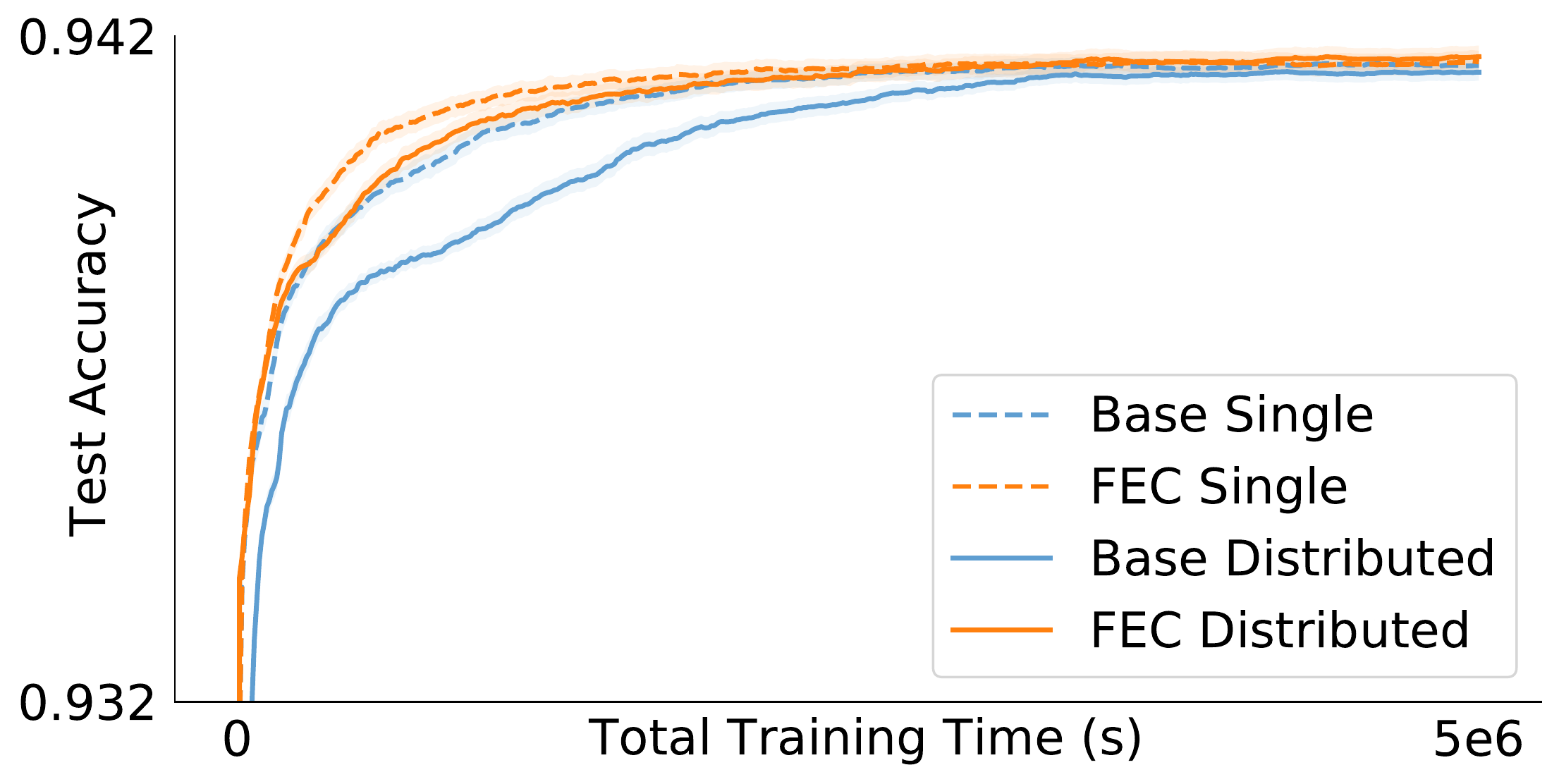}
    
\caption{Comparison of FEC and baseline in single-threaded (serial) and distributed  (parallel) search infrastructure, using the NAS-Bench-101 setup. FEC shows improvements in both cases. Lines are the mean across 500 experiments with the shading being $\pm$ 2 SEM.}
\label{fec_distributed_fig}
\end{figure}

We can compare baseline and FEC experiments side-by-side on NAS-Bench-101, where serial experiments can make progress in a reasonable amount of time. Distributed search does affect the dynamics, as evidenced by the difference between the single and distributed lines of Figure~\ref{fec_distributed_fig}. Still, as the figure shows, FEC provides a speedup in both the serial and the distributed settings.

\subsubsection{Hash Collisions} 
\label{hash_collisions_sec}

\begin{figure*}[ht]
\begin{subfigure}{.4\textwidth}
    \centering
   \includegraphics[width=1.0\linewidth,trim={10mm 11mm 0 0},clip]{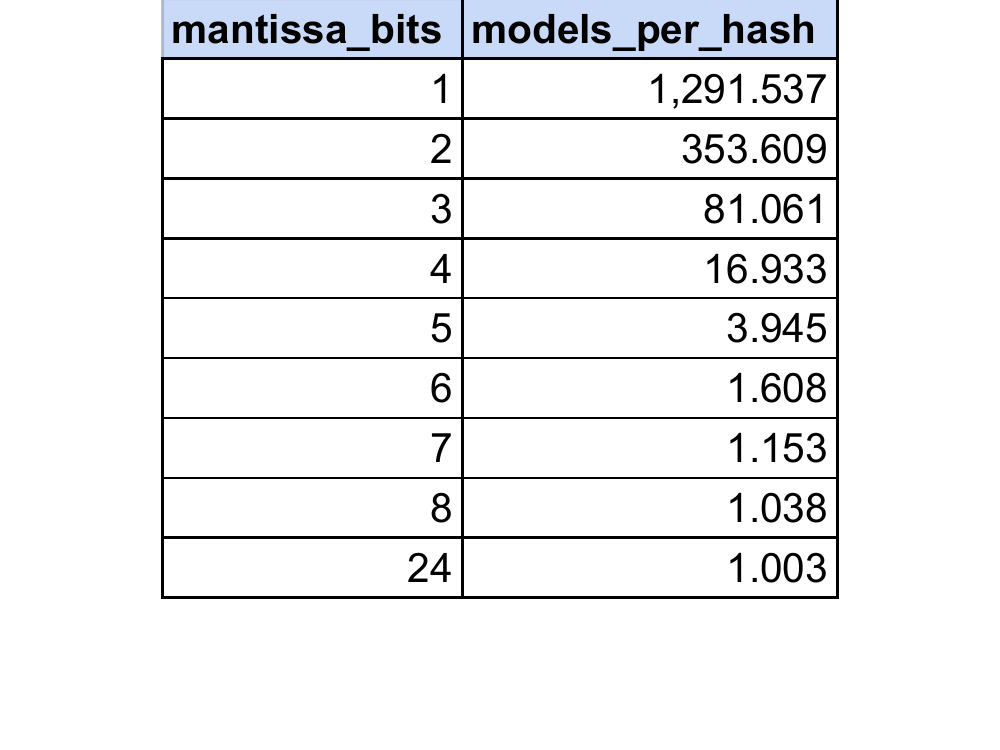}
\end{subfigure}
\hspace{5mm}
\begin{subfigure}{.55\textwidth}
    \centering
    \includegraphics[width=1.0\linewidth]{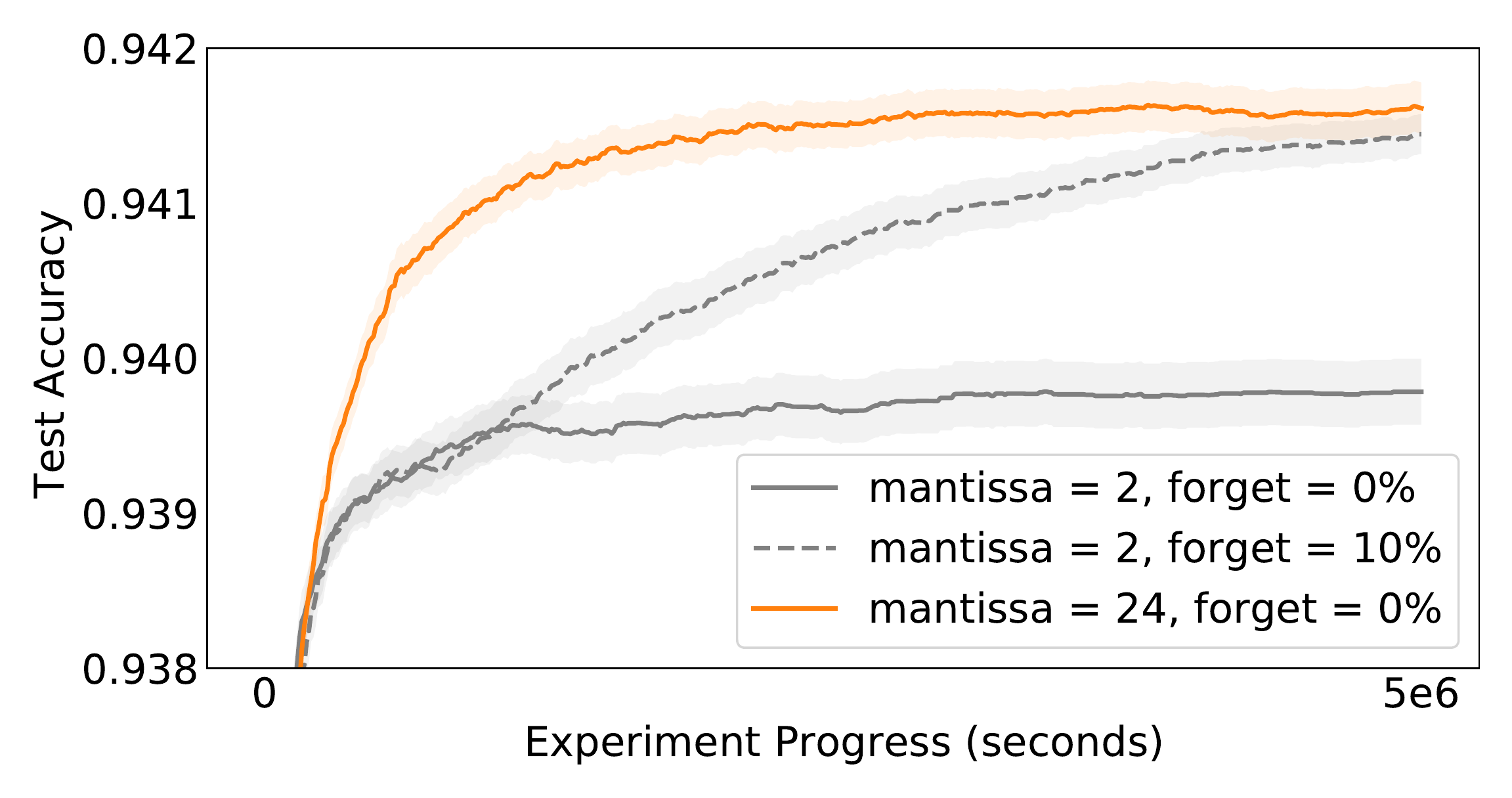}
\end{subfigure}
\caption{Hash Collisions in NAS-Bench-101. LEFT: Average number of models per UFH as a function of mantissa bits. RIGHT: Search progress for different mantissa bits and amounts of forgetfulness. Lines are the mean across experiments with the shading being $\pm$ 2 SEM. Note that large amounts of collisions are necessary to cause degradation. Degradation recovery happens with cache forgetfulness.}
\label{fec_collisions_fig}
\end{figure*}

In our experience, hash collisions were present in all setups but they were infrequent. The results of Section~\ref{results_sec} did not require tuning hash parameters, but nevertheless we consider this issue in more detail here using the NAS-Bench-101 setup. We repeat all the experiments with different hash precisions. That is, we reduce the mantissa bits ($m_{bits}$) parameter in order to force collisions (Figure~\ref{fec_collisions_fig}--LEFT). Figure~\ref{fec_collisions_fig}--RIGHT's solid gray line shows that the hashes must be very imprecise for the search quality to be seriously affected.

In cases where collisions are frequent enough to affect the search, it is possible that their impact can be ameliorated by a ``forgetful'' cache, as follows. Collisions affect the search by poisoning the cache for a specific key. In our FEC implementation thus far, the poisoned key remains in the cache forever. To address this, we might let the cache forget the key by having it be removed with a small probability upon being queried. This can result in a recovery from the search degradation caused by the collisions as shown by (Figure~\ref{fec_collisions_fig}--RIGHT)'s dashed gray line. We used a fixed probability of $0.1$, but note that it may make sense to forget more frequently during the first few queries of a given key and less later on (\eg forgetting could have a probability of $1/N$ for a key that has been queried $N$ times since it was last added to the cache). This forgetful FEC is shown in Method~\ref{alg:fec_forget}. Further research might clarify the generality of this approach.

\subsubsection{Evaluation Noise} 
\label{eval_noise_sec}

Evaluation noise can also poison the cache. A high statistical outlier might bias the search toward its region of the search space, as it will unfairly have more descendents. A low statistical outlier may represent a lost opportunity, as it may die before it has any offspring. Both of these negative effects are boosted by caching, as the outliers may remain in the cache forever.

\begin{figure}[ht]
\begin{subfigure}{.48\linewidth}
    \centering
    \includegraphics[width=\linewidth]{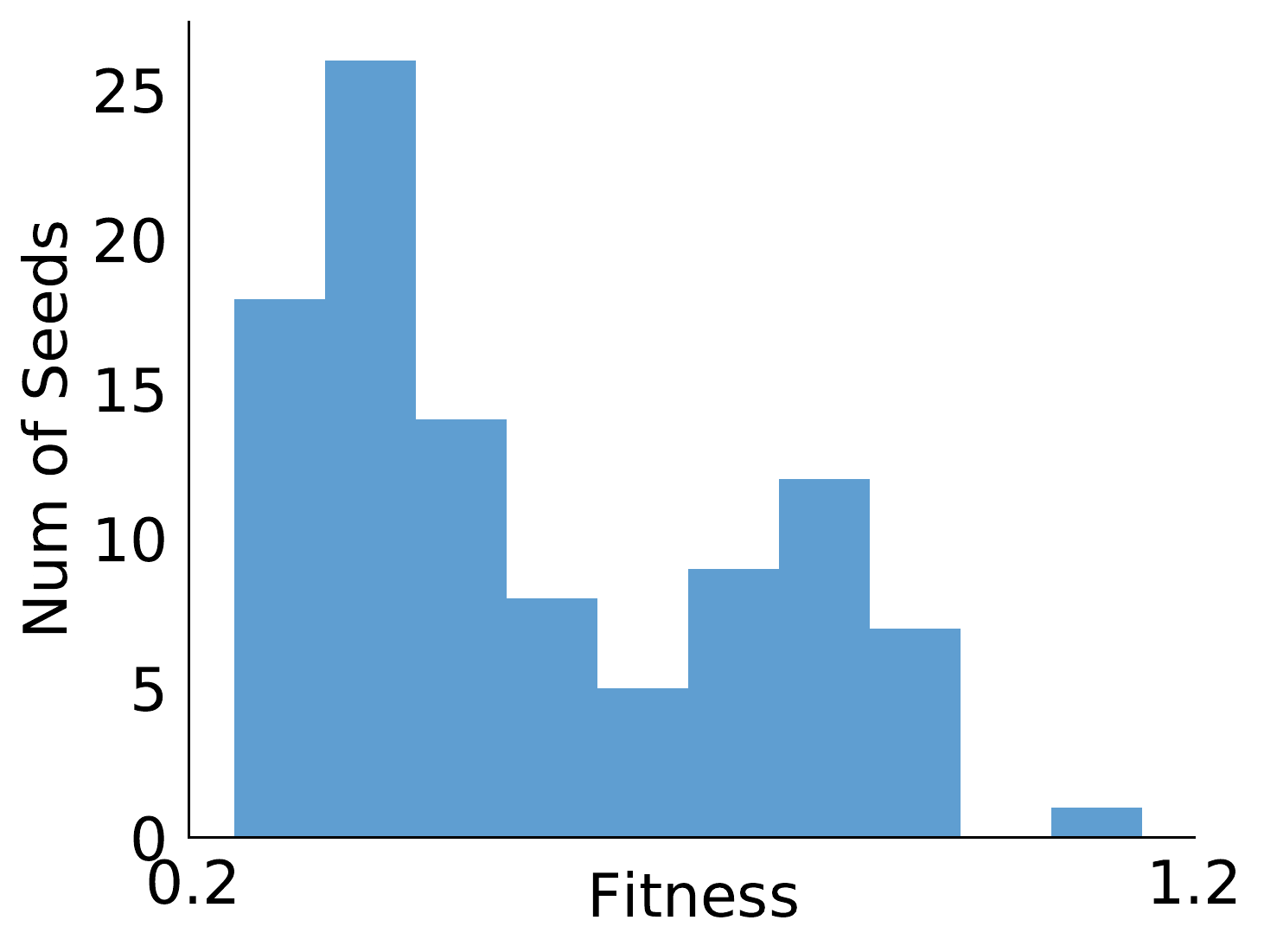}
\end{subfigure}
\begin{subfigure}{.48\linewidth}
    \centering
    \includegraphics[width=\linewidth]{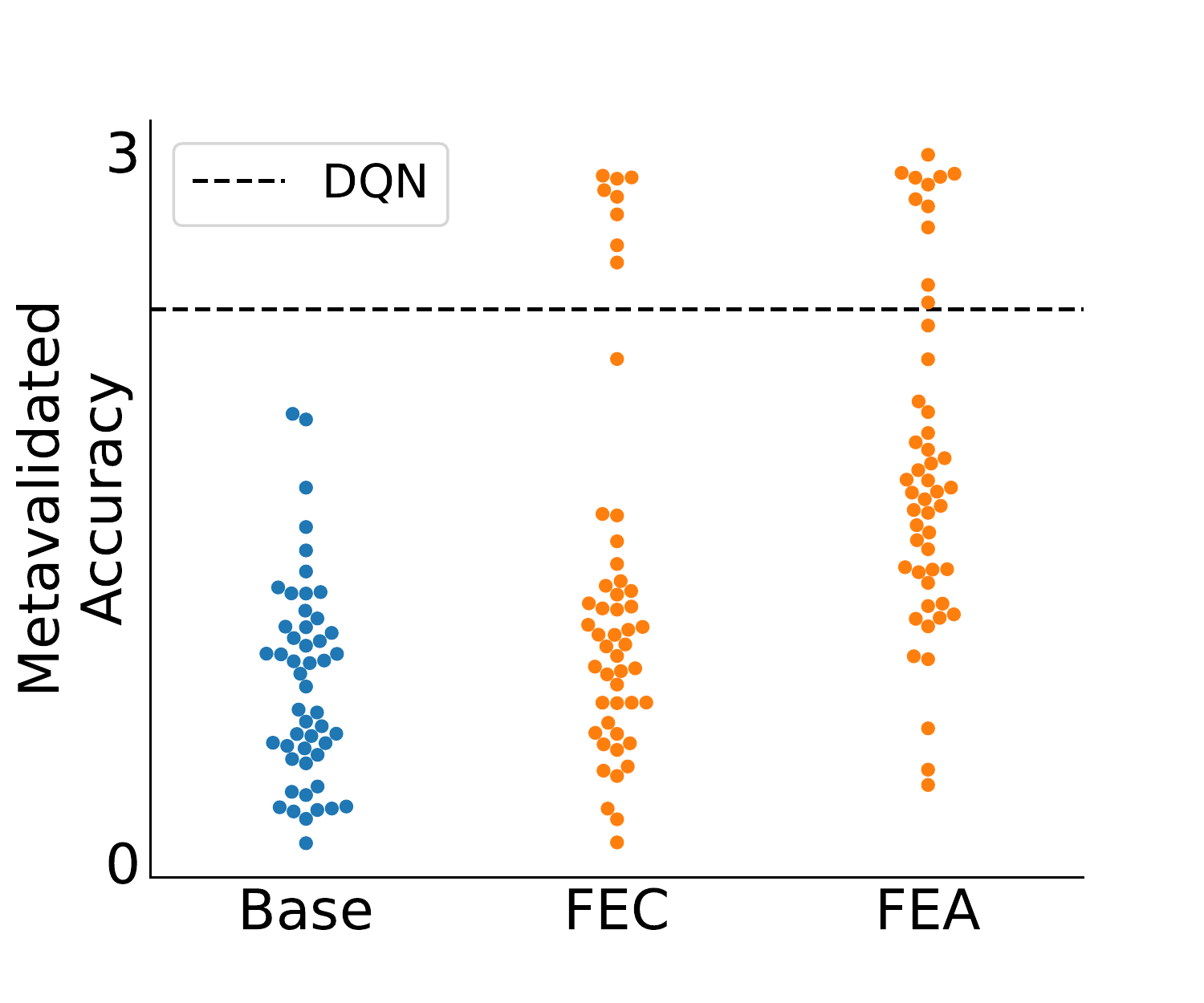}
\end{subfigure}
\caption{FEC and evaluation noise. LEFT: histogram of the fitnesses of a many evaluations of a single discovered algorithm, showing the large amounts of noise in AutoRL training. RIGHT: Comparison of search experiment results without FEC, with FEC, and with FEA.}
\label{metavalid_aurotl_agg_fig}
\end{figure}

Moderate amounts of noise have not prevented FEC from providing benefits, however. One of our setups, AutoRL, has a large amount of evaluation noise, due to the inherently high stochasticity of reinforcement learning training during candidate evaluation (Figure~\ref{metavalid_aurotl_agg_fig}--LEFT). Nevertheless, the method remained effective in the AutoRL experiments of Section~\ref{results_sec}.

If the goal of the search is to discover algorithms with high \emph{mean fitness}, then it may be possible to use a different form of caching. In FEC, we cached the candidate's fitness. Thus, during the experiment, a given candidate's fitness is fully evaluated only once, and it is retrieved each time its corresponding UFH is seen after that. As we have shown, that results in a speed improvement. Instead, we could opt to use the hashes to combat noise. Namely, we could consider storing a running mean of the evaluations for each functional hash. As equivalent candidates are sampled, their fitnesses are aggregated in a dictionary keyed by the hash. This aggregated fitness is the one that we use for evolutionary selection. We can refer to this method as \emph{functional equivalence aggregation} (FEA) as seen in Method~\ref{alg:fea}. Thus, candidates that are seen more frequently will end up with more accurate evaluations. More investigation is required, but preliminary results do show an improvement in the bulk of the distribution of the reevaluated accuracies of the best evolved candidates (Figure~\ref{metavalid_aurotl_agg_fig}--RIGHT).

\subsection{Structural \vs Functional Equivalence}
\label{struct_func_equiv_sec}

\begin{figure}[ht]
    \centering
\includegraphics[width=0.9\linewidth]{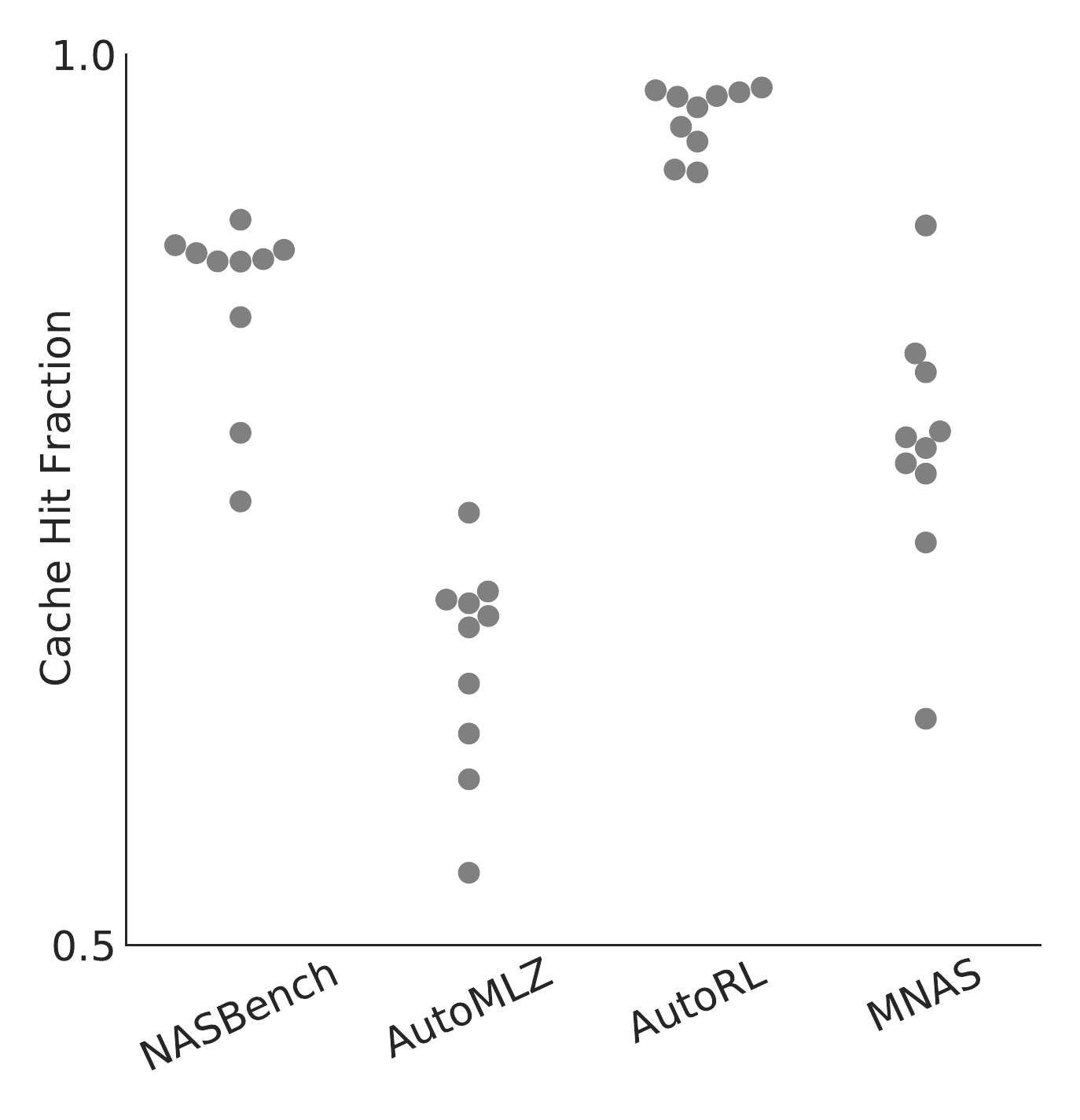}
\caption{Cache Hit Fraction. Each point corresponds to an FEC experiment and indicates the fraction of candidates that were retrieved from the cache. \Eg A ``0.8'' means that 80\% of proposed candidates had been proposed before. Note that all experiments are above 0.5, indicating the propensity of evolutionary methods to repeat candidates.}
\label{repeat_counts_fig}
\end{figure}

\begin{figure*}[ht]
\begin{subfigure}{0.5\textwidth}
    \centering
    \includegraphics[width=0.98\linewidth]{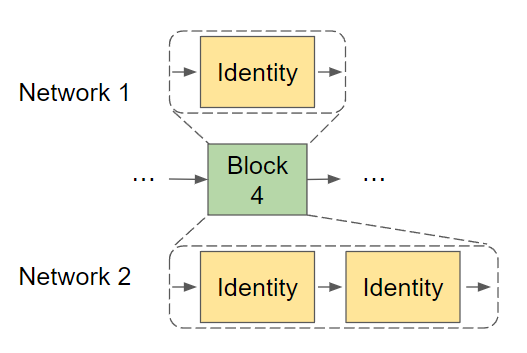}
\end{subfigure}
\begin{subfigure}{0.24\textwidth}
    \begin{code}{1.0\textwidth}{1.7in}{codebackground}
    \codeline{\codedef{def} InitializePass():}
      \codeline{\codetab s2 = 0.01}
    \codeskip
    \codeline{\codedef{def} ForwardPass():}
      \codeline{\codetab s1 = s1 + s4}
      \codeline{\codetab s5 = dot(v1, v0)}
      \codeline{\codetab s1 = s1 + s5}
    \codeskip
    \codeline{\codedef{def} BackwardPass():}
      \codeline{\codetab s3 = s0 - s1}
      \codeline{\codetab s3 = s2 * s3}
      \codeline{\codetab s4 = s4 + s3}
      \codeline{\codetab v2 = s3 * v0}
      \codeline{\codetab v1 = v1 + v2}
    \end{code}
\end{subfigure}
\begin{subfigure}{0.24\textwidth}
    \begin{code}{1.0\textwidth}{2.3in}{codebackground}
    \codeline{\codedef{def} InitializePass():}
      \codeline{\codetab s2 = -0.0124237}
    \codeskip
    \codeline{\codedef{def} ForwardPass():}
      \codeline{\codetab s1 = s1 + s4}
      \codeline{\codetab s5 = dot(v1, v0)}
      \codeline{\codetab s1 = s1 + s5}
    \codeskip
    \codeline{\codedef{def} BackwardPass():}
      \codeline{\codetab \codecomment{3 ineffective lines:}}
      \codeline{\codetab v2 = s2 * v1}
      \codeline{\codetab v2 = s4 * v1}
      \codeline{\codetab v2 = s4 * v0}
      \codeline{\codetab s3 = s2 * s3}
      \codeline{\codetab s4 = s3 + s4}
      \codeline{\codetab \codecomment{v2 gets overwritten:}}
      \codeline{\codetab v2 = s3 * v0}
      \codeline{\codetab v2 = s3 * v0} 
      \codeline{\codetab v1 = v1 + v2}
    \end{code}
\end{subfigure}
\caption{Examples of candidates that are functionally equivalent but structurally different. LEFT: Two MNAS neural networks. Both use an identity op for the fourth block, and so the different number of layer repetitions does not change their functionalities. See Figure~\ref{mnas_setup_fig} for full architecture. RIGHT: Two AutoML-Zero programs. Both perform affine regression but the longer program contains ineffective operations.}
\label{functional_structural_fig}
\end{figure*}

UFH is an effective method for handling the many repeated candidates proposed by evolutionary algorithms (Figure~\ref{repeat_counts_fig}). The reasons for all these repeats vary between search spaces. For example, AutoRL represents functions as fixed-size compute graphs. Multiple graphs can represent the same function in various ways: isomorphic graphs, non-functional vertices (not on the path from input to output), irrelevant operations (\eg adding 0). Examples from actual experiments can be seen in Figure~\ref{functional_structural_fig}. While UFH is not the only method for detecting repeats, the rest of this section will argue that it is simpler and more general than the alternative structural---as opposed to functional---approaches.

Detecting equivalent candidates based on their structure can be difficult, as can be seen from the examples in Figure~\ref{functional_structural_fig}. For instance, comparing two compute graphs for isomorphisms is computationally hard, as it requires figuring out corresponding vertices. Even if that were possible due to small graph sizes, the problem remains that some equivalent graphs are not structurally isomorphic. This is the case in the addition-of-zero example (note that ``zero'' can take complex forms such as $x^2 - x * x$). When the candidates are expressed as code, even more difficulties arise. In AutoML-Zero, for example, operations can be rendered semantically ineffective if their outputs are overwritten in the memory before being read. Instead of requiring complex comparisons between graphs or between pieces of code, UFH simply relies on the candidate's execution---it's reaction to data.

In addition, UFH is conceptually the same for all search spaces. Given a codebase that can already evaluate candidates (which it must), introducing UFH amounts to a small alteration of the evaluation routine, as indicated by the highlighted lines in Method~\ref{hash_alg}. Structural equivalence is not as general. The set of steps required to compare two pieces of code for structural equivalence is not the same as those required to compare two compute graphs. Moreover, the details of the search space matter too (\eg are the operations commutative?) Functional equivalence automatically handles all of this under a unified framework.


\subsection{Optional Use of Fake Data}
\label{discussion_fake_data_sec}

One of the key components of the UFH method is to fix a small set of canonical training and validation examples. For most of this paper, we randomly selected them from the dataset of the experimental setup in question. For example, in MNAS, the hashing examples came from the CIFAR-10 dataset. Let us refer to this practice as hashing with \emph{real data}. As it turns out, real data is not necessary. While hashing, it is possible to use \emph{any} data, even examples that make no sense, such as those with mismatched inputs and labels. The goal of the hashing examples is to activate all relevant parts of the candidate being hashed, especially those that matter when parsing the full dataset. For this reason, enough examples of real data are the ideal choice, as it comes from the desired distribution. The use of fake data may not fully probe the candidate and therefore lead to more hash collisions. Nevertheless, there are circumstances where we may want to use fake data anyway, such as when real data is computationally expensive. This was the case in our AutoRL setup, where the data arises from physics simulations. In addition to compute costs, collecting representative samples in an RL environment is highly nontrivial, as it requires policies that can sufficiently explore the environment. Therefore, in this paper we used a fixed set of randomly sampled states, rewards, and actions as fake data for all the AutoRL experiments.

\subsection{Hash Construction Guidelines}
\label{hash_construction_sec}

UFH hashing has three parameters that are relatively easy to set: the number of mantissa bits ($m_{bits}$), the number of examples ($N_E$), and the number of random seeds ($N_S$). In this section, we describe how to choose their values and how to verify the appropriateness of the choice.

The $m_{bits}$ should not exceed the precision of the computations done by the model. In particular, the $m_{bits}$ cannot be higher than 52 for the \textcode{float64} type, 23 for \textcode{float32}, 10 for \textcode{float16}, and 7 for \textcode{bfloat16}. Setting the maximum value, however, should be avoided because roundoff errors will effectively reduce the precision of the hashable outputs, which can in turn incur avoidable cache misses. Practitioners often have a sense of what is a meaningful precision in their use case; otherwise, half the maximum precision can be used as a starting guess. This lack of care in choosing the $m_{bits}$ is justified because a small increase in $N_E$ and $N_S$ will easily compensate for a low $m_{bits}$, as we explain next.

The $N_E$ and $N_S$ directly affect the compute cost of hashing because the total number of iterations in the evaluation scales as $N_E \times N_S$ for each candidate. Thus, the cost of an experiment with FEC will scale as $K \times N_E \times N_S + K \times (1-h) \times N_T$, where $K$ is the total number of candidates in the experiment, $h$ is the cache hit rate, and $N_T$ is the number of training examples used when the cache is missed. In contrast, without FEC, the cost scales as just $K \times N_T$. Thus, for hashing to provide significant savings, we need that $N_E \times N_S \ll h \times N_T$. From Figure~\ref{repeat_counts_fig}, we see that $h$ is high enough that, for order-of-magnitude estimates, we only require that\\
\centerline{$N_E \times N_S \ll N_T$}\\
Noting that the number of training steps $N_T$ in machine learning tends to be large, we are left with a lot of flexibility. From our experiments, the empirical speed gains were in the range 2x--10x, but never as large as 100x. Based on this, we can decide that we do not care to set these hyperparameters to allow for 100x gains, so we could fix $N_E = N_S = \sqrt{N_T} / 10$ as a starting point. Of course, we want $N_E \gg 1$, as training effects on function may have an effect only after a few steps of training have occurred (in an extreme case for example, in AutoML-Zero, an operation may not have any effect on the output until $N_E$ reaches the length of the program). These heuristic estimates are only intended to give a sense of why our method does \emph{not} require the empirical tuning of the hyperparemeters it introduces. In practice, we typically used $N_E \approx 10$ and $N_S \approx 3$ and that proved sufficient.

Once the parameters have been set, it can be worth verifying their appropriateness without having to do full-size controlled experiments. To do this, we can run a short \emph{FEC counterfactual} experiment, designed to verify that the collision rate is sufficiently low. Such an experiment shares the same setup as a regular evolutionary search experiment with FEC, except that upon a cache hit we still do a full model evaluation. Then we compare the evaluation result with the result retrieved from the cache. If the difference between the two is larger than a predefined roundoff error, we consider such an event to be a cache collision. The hyperparameters ($m_{bits}, N_E, N_S$) should minimize the cache collision rate without significantly degrading the cache hit rate. In practice, the main use of FEC counterfactual experiments is to detect egregious collision rates that may point to bugs in the code.



\section*{Author Contributions}

ER conceived the project. ER, CL, DS, and QVL produced the first implementation and demonstrated plausibility. RG, SJ, MM, YM, CdS, CL, DS, and ER wrote the code. RG, SJ, MM, and YM carried out the main experiments and analysis. ER wrote the paper, with contributions from RG. JD and CdS prepared the figures. CdS open-sourced the code. ER and RG coordinated the work. ER supervised.

\section*{Acknowledgements}

We would like to thank Aman Agrawal for code contributions; Charles Weill for experimental work; and Mitchell McIntire, JD Co-Reyes, and Aleksandra Faust for helpful discussions, as well as the larger Google Brain team.


\bibliography{ms}
\bibliographystyle{plain}



\supplheader{}

\supplsection{FCM Technique Details}{fcm_details}{
\label{suppl_fcm_sec}

Analogously to Method~\ref{alg:fec} in the main text, Method~\ref{alg:fcm} shows the use of FCM in the context of regularized evolution. As the highlighted blue text indicates, the child is mutated until its hash differs from the parent's hash. In practice, to avoid infinite loops, there is often a maximum number of trials; if reached, the child is returned as is.

\begin{algorithm}
\caption{Regularized Evolution \alghlight{with FCM (\alghlightcolor)}}
\label{alg:fcm}
\small
\begin{algorithmic}
\STATE $population \gets$ empty queue
\STATE $n = 0$  \algspacedcomment{Number of sampled candidates.}
\STATE \algcomment{P is the population size meta-parameter.}
\WHILE{$|population| < P$}
    \STATE $seed \gets$ RandomCandidate()
    \STATE $seed.fitness \gets$ Evaluate($seed$)
    \STATE $population.enqueue(seed)$
    \STATE $n = n + 1$
\ENDWHILE
\STATE \algcomment{N is the total number of candidates meta-parameter.}
\WHILE{n < N}
    \STATE \algcomment{T is the tournament size meta-parameter.}
    \STATE $tournament \gets$ RandomSubset($population$, size=$T$)
    \STATE $parent \gets $ CandidateWithBestFitness($tournament$)
    \STATE $child \gets$ Mutate($parent$)
    \color{\alghlightcolor}
    \STATE $parent.hash$ = UnifiedFunctionalHash($parent$)
    \WHILE{UnifiedFunctionalHash($child$) == $parent.hash$}
        \STATE $child \gets$ Mutate($child$)
    \ENDWHILE
    \color{black}
    \STATE $child.fitness \gets$ Evaluate($child$)
    \STATE $population.enqueue(child)$
    \STATE $population.dequeue()$ \algspacedcomment{Remove oldest.}
    \STATE $n = n + 1$
\ENDWHILE
\STATE \textbf{Return} CandidateWithBestFitness($population$)
\end{algorithmic}
\end{algorithm}

}  

\supplsection{Tabulist Technique Details}{tabulist_details}{
\label{suppl_tabulist_sec}

Analogously to Method~\ref{alg:fec} in the main text, Method~\ref{alg:tabulist} shows the use of the tabulist technique in the context of regularized evolution. ``\textcode{tabulist[key]}'' counts the number of occurrences of the functional hash ``\textcode{key}'' during the experiment. The final \textcode{while} loop prevents children with hashes that have been seen more than \textcode{K} times from being introduced into the population.


\begin{algorithm}
\caption{Regularized Evolution \alghlight{with tabulist (\alghlightcolor)}}
\label{alg:tabulist}
\small
\begin{algorithmic}
\STATE $population \gets$ empty queue
\STATE \alghlight{$tabulist \gets $ empty hash table, default value 0} \algspacedcomment{Candidate hash to seen count.}
\STATE $n = 0$  \algspacedcomment{Number of sampled candidates.}
\STATE \algcomment{P is the population size meta-parameter.}
\WHILE{$|population| < P$}
    \STATE $seed \gets$ RandomCandidate()
    \color{\alghlightcolor}
    \STATE $key \gets $ UnifiedFunctionalHash($seed$)
    \STATE $tabulist[key] \gets$  $tabulist[key]$ + 1
    \color{black}
    \STATE $seed.fitness \gets$ Evaluate($seed$)
    \STATE $population.enqueue(seed)$
    \STATE $n = n + 1$
\ENDWHILE
\STATE \algcomment{N is the total number of candidates meta-parameter.}
\WHILE{n < N}
    \STATE \algcomment{T is the tournament size meta-parameter.}
    \STATE $tournament \gets$ RandomSubset($population$, size=$T$)
    \STATE $parent \gets $ CandidateWithBestFitness($tournament$)
    \STATE $child \gets$ Mutate($parent$)
    \color{\alghlightcolor}
    \STATE $key \gets $ UnifiedFunctionalHash($seed$)
    \STATE \algcomment{K is the tabulist max seen count meta-parameter.}
    \WHILE{$tabulist[key] \geq K$}
        \STATE $child \gets$ Mutate($child$)
        \STATE $key \gets $ UnifiedFunctionalHash($child$)
    \ENDWHILE
    \STATE $tabulist[key] \gets$  $tabulist[key]$ + 1
    \color{black}
    \STATE $child.fitness \gets$ Evaluate($child$)
    \STATE $population.enqueue(child)$
    \STATE $population.dequeue()$ \algspacedcomment{Remove oldest.}
    \STATE $n = n + 1$
\ENDWHILE
\STATE \textbf{Return} CandidateWithBestFitness($population$)
\end{algorithmic}
\end{algorithm}

}  

\supplsection{Time-course of NAS-Bench-101 Expts.}{nasbench_timecourses}{

Figure~\ref{nasbench_valid_fig} shows the time-course plots of the validation fitnesses for the three techniques on the NAS-Bench-101 setup. Validation time-courses are monotonically increasing because they show the running maximum. Corresponding test fitnesses can be seen in Figure~\ref{nasbench_test_fig}.


\begin{figure}[!t]
    \begin{centering}
        \includegraphics[width=0.32\columnwidth]{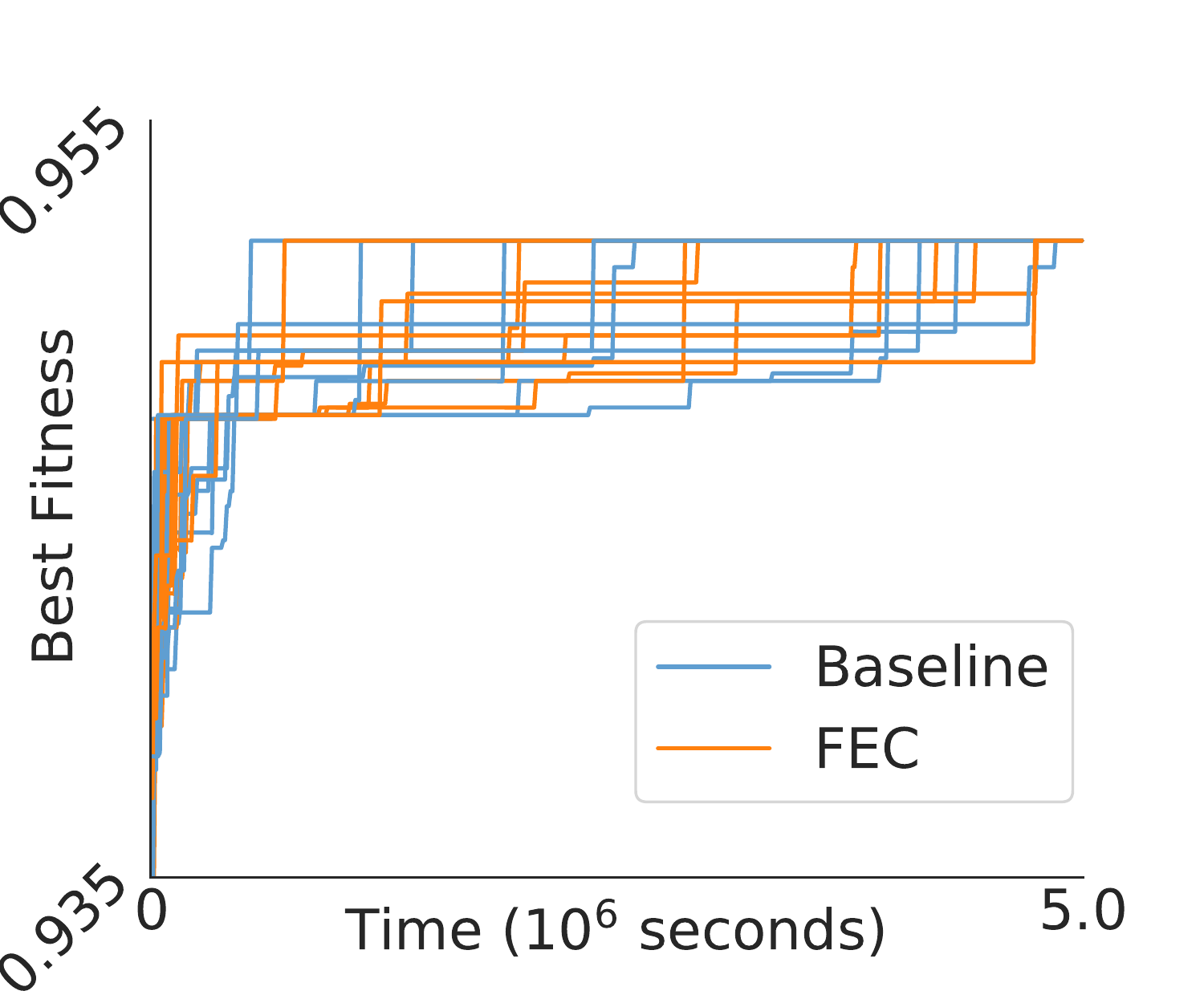}
        \includegraphics[width=0.32\columnwidth]{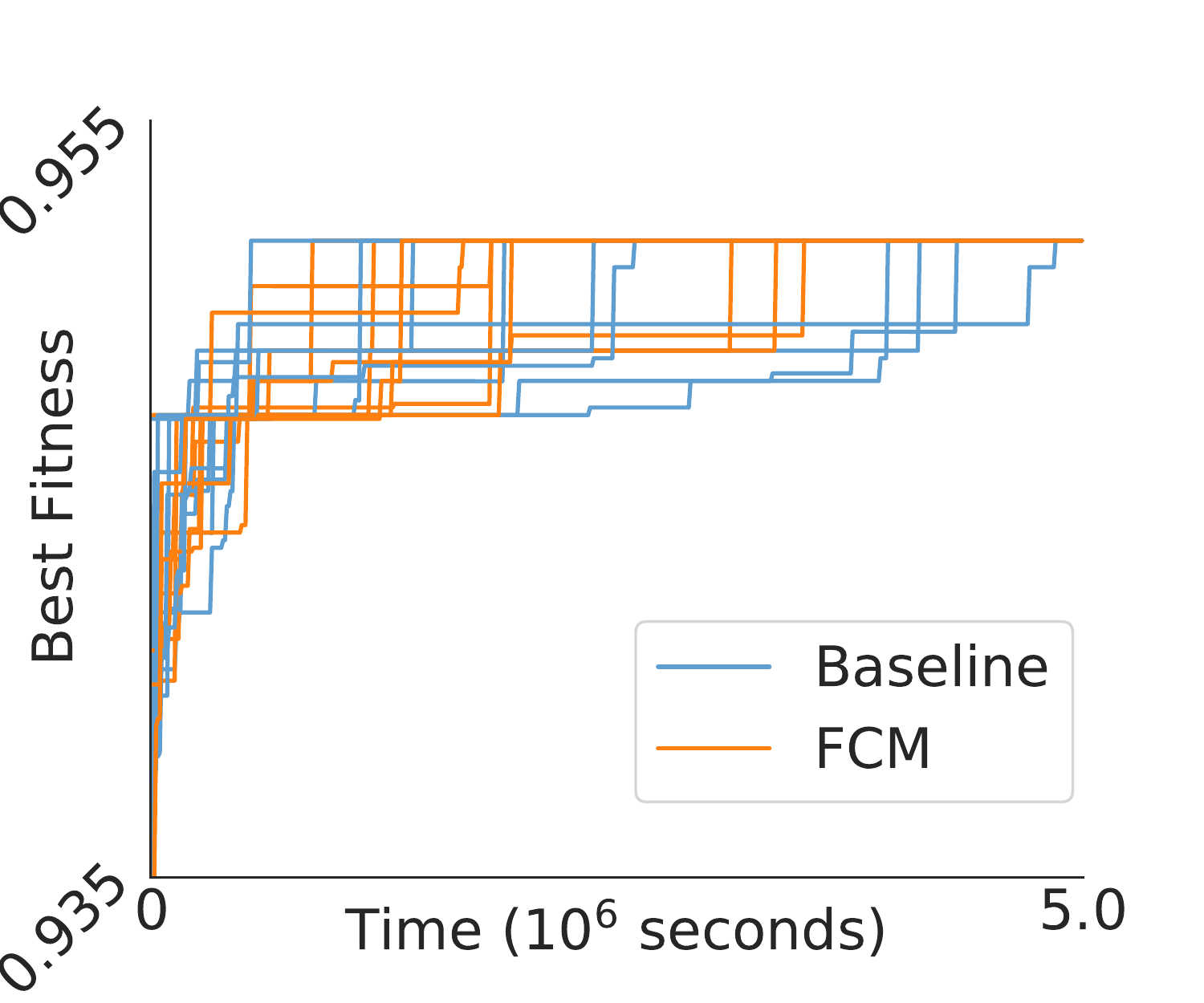}
        \includegraphics[width=0.32\columnwidth]{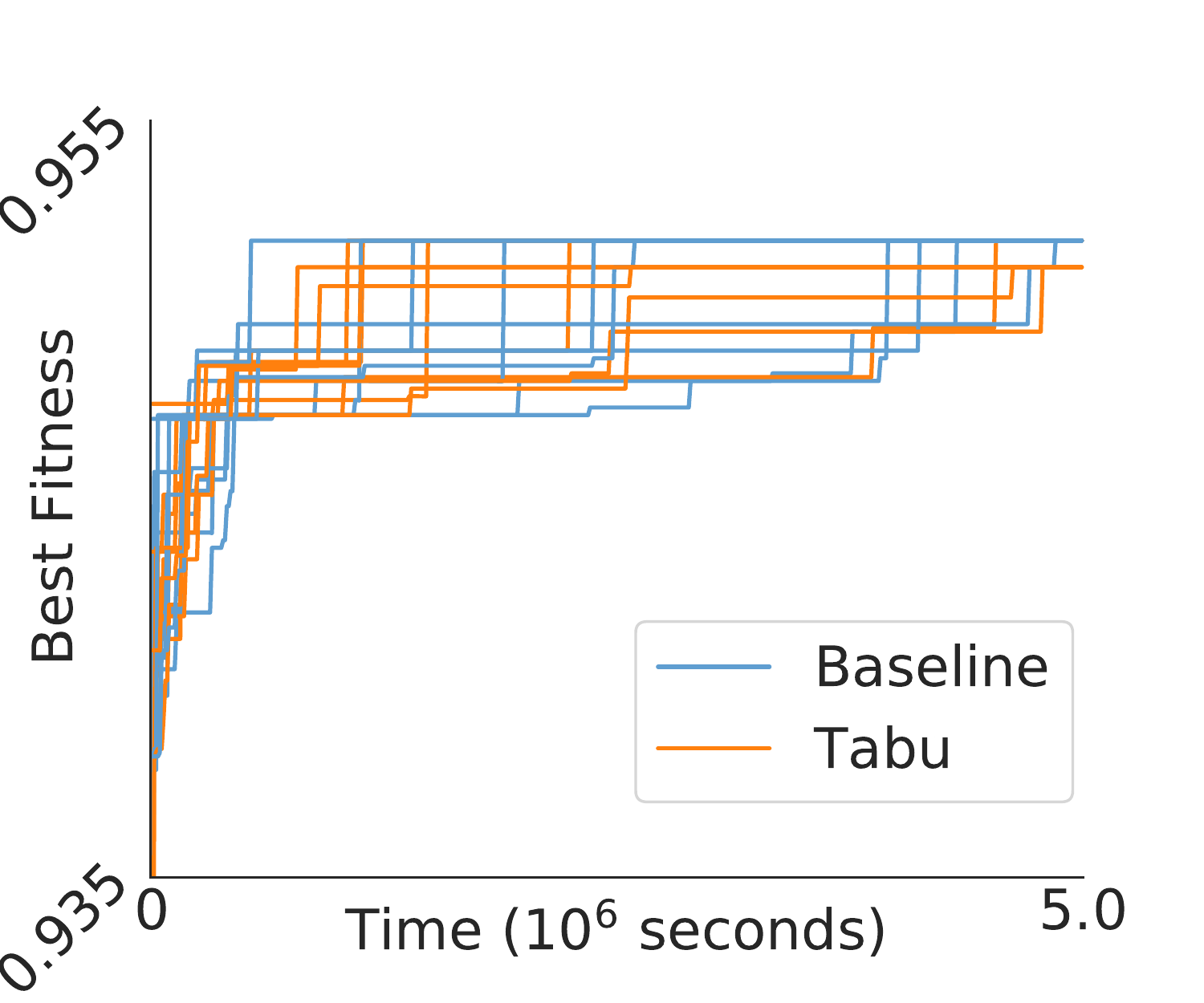}
        \caption{NAS-Bench-101 validation fitness time-courses. LEFT: FEC, MIDDLE: FCM, and RIGHT: tabulist.}
        \vspace{-2pt}
        \label{nasbench_valid_fig}
    \end{centering}
\end{figure}

\begin{figure}[!t]
    \begin{centering}
        \includegraphics[width=0.32\columnwidth]{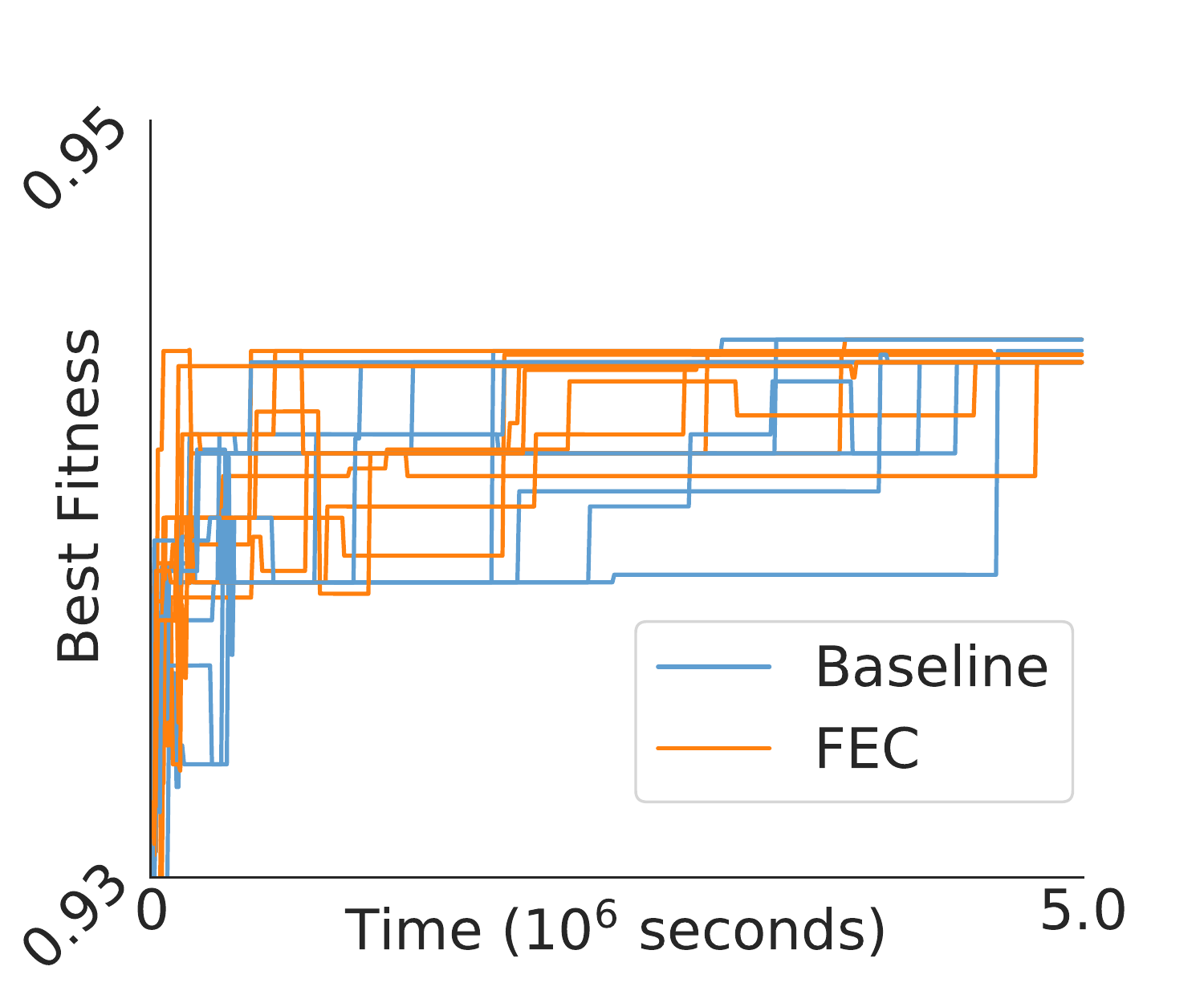}
        \includegraphics[width=0.32\columnwidth]{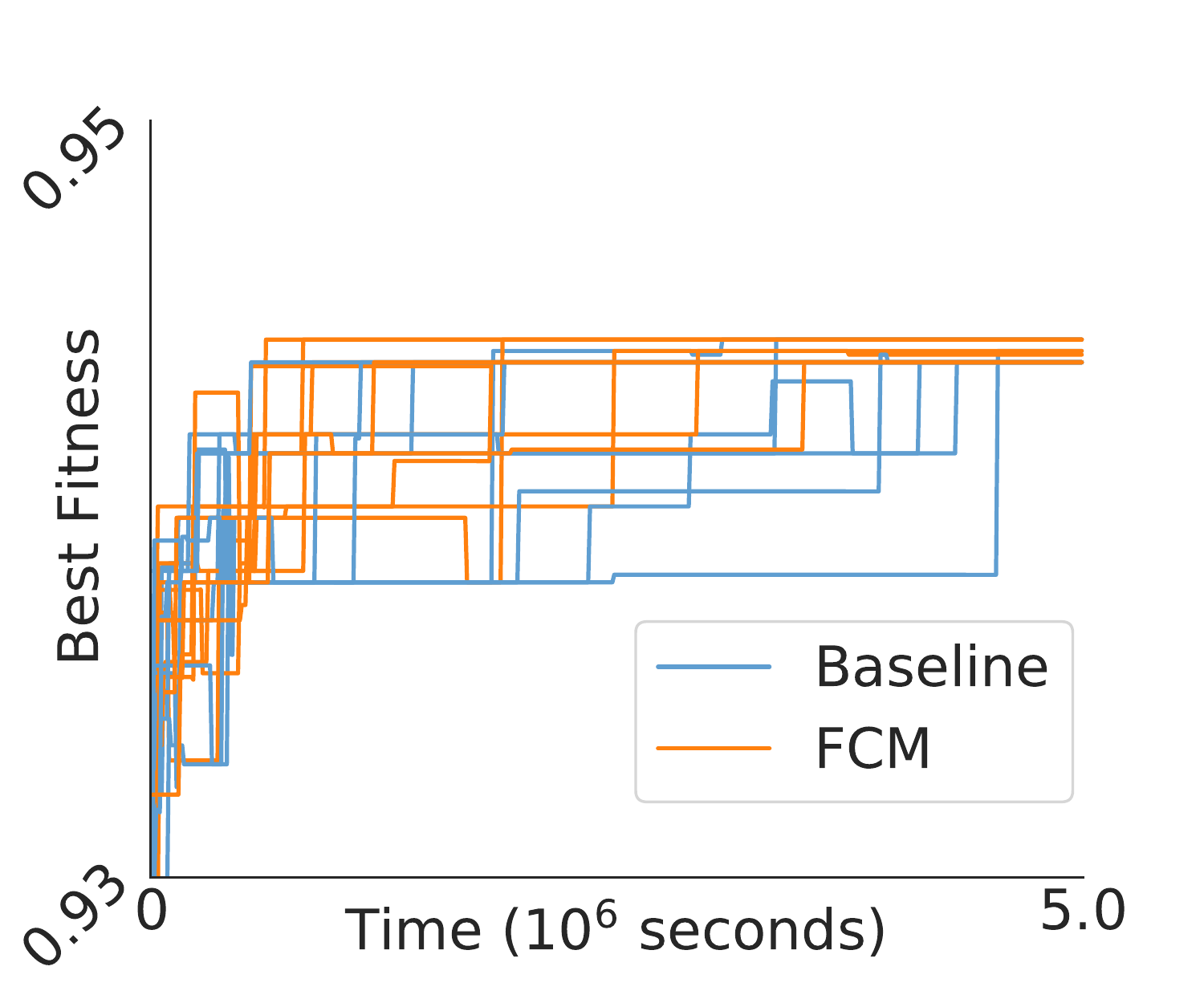}
        \includegraphics[width=0.32\columnwidth]{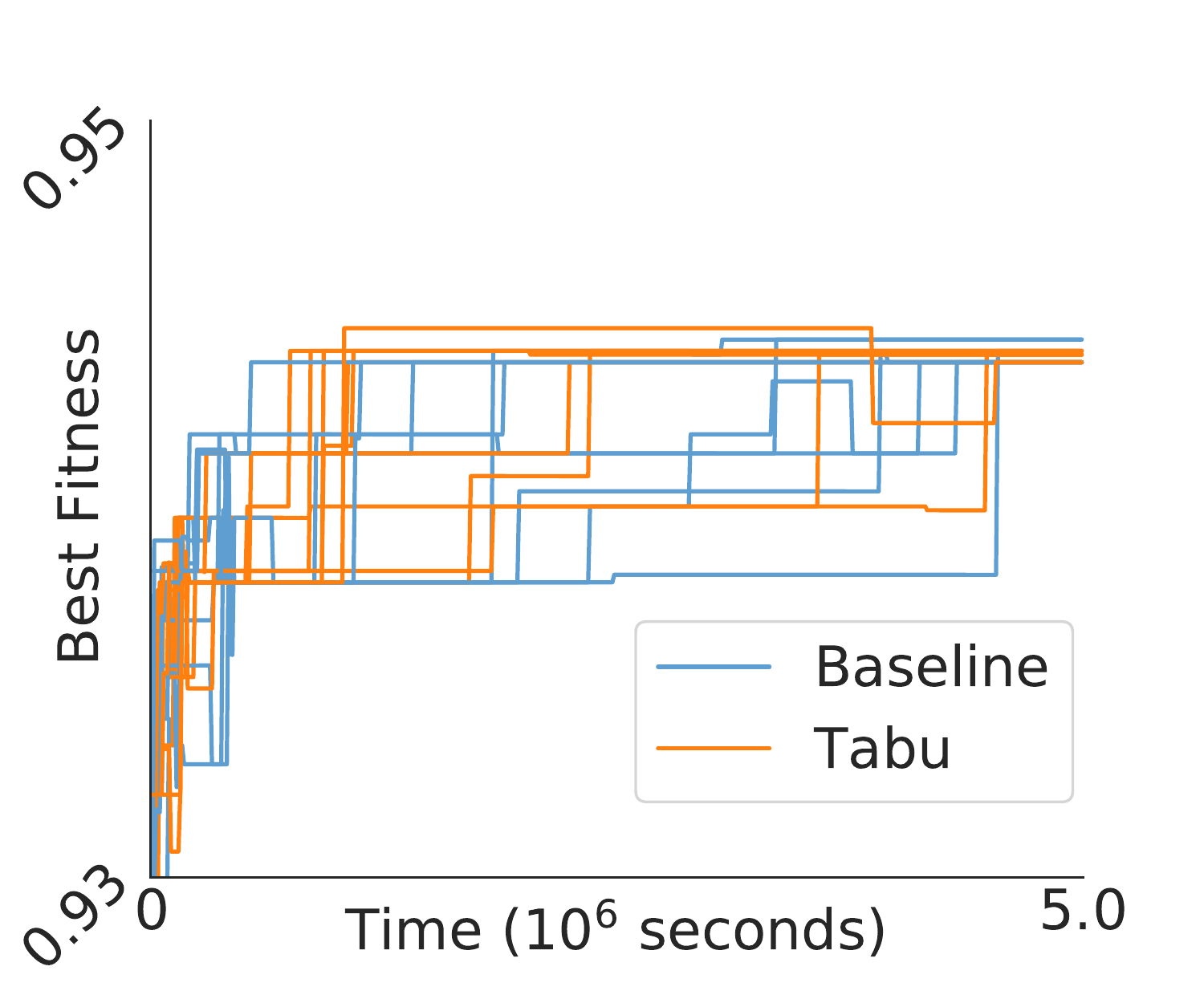}
        \caption{NAS-Bench-101 test fitness time-courses. LEFT: FEC, MIDDLE: FCM, and RIGHT: tabulist.}
        \vspace{-2pt}
        \label{nasbench_test_fig}
    \end{centering}
\end{figure}

}  

\supplsection{Time-course of AutoML-Zero Expts.}{amlz_timecourses}{

Figure~\ref{amlz_suppl_fig} shows the time-course plots for the three techniques on the AutoML-Zero setup.


\begin{figure}[!t]
    \begin{centering}
        \includegraphics[width=0.32\columnwidth]{AMLZ_curves.pdf}
        \includegraphics[width=0.32\columnwidth]{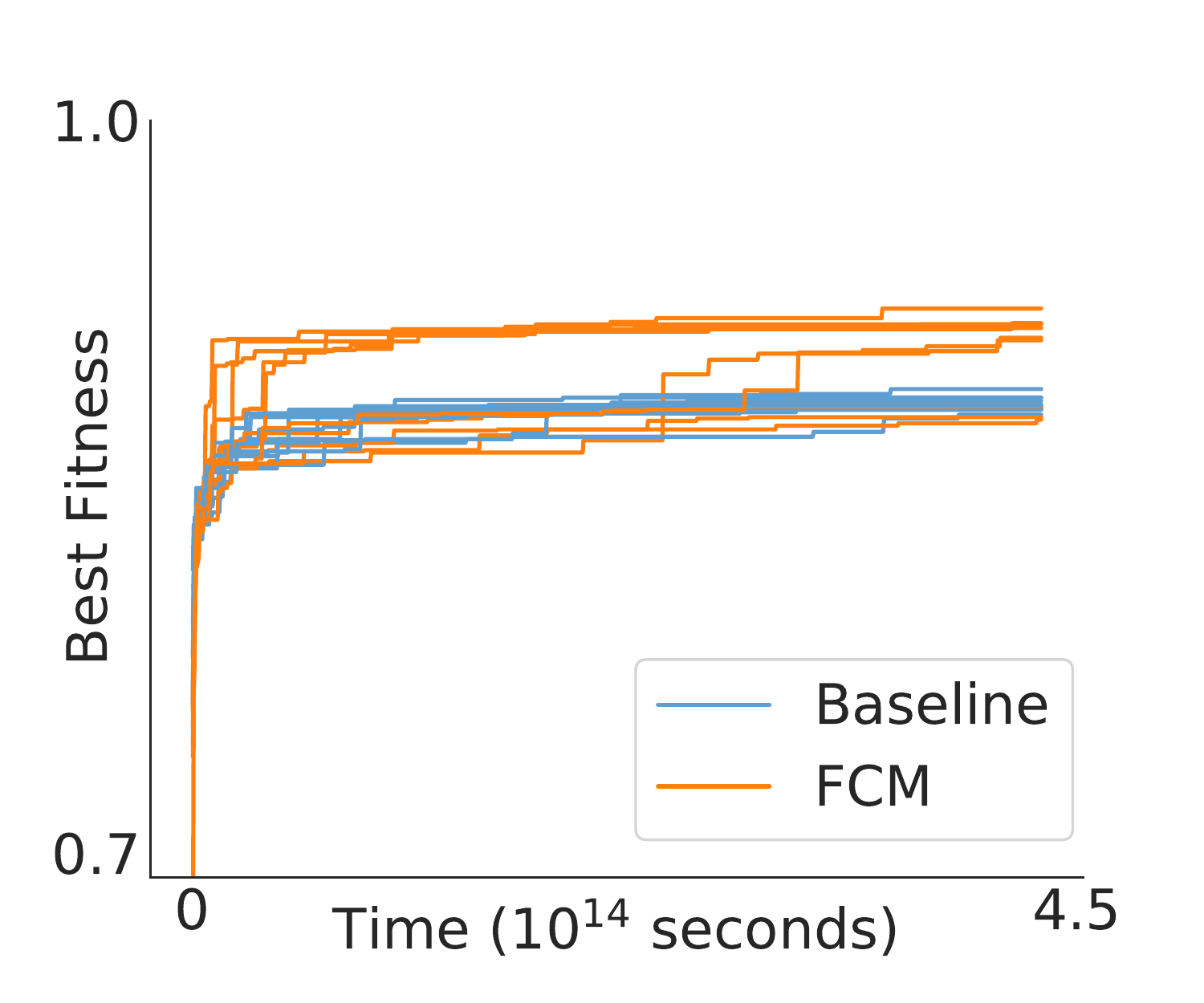}
        \includegraphics[width=0.32\columnwidth]{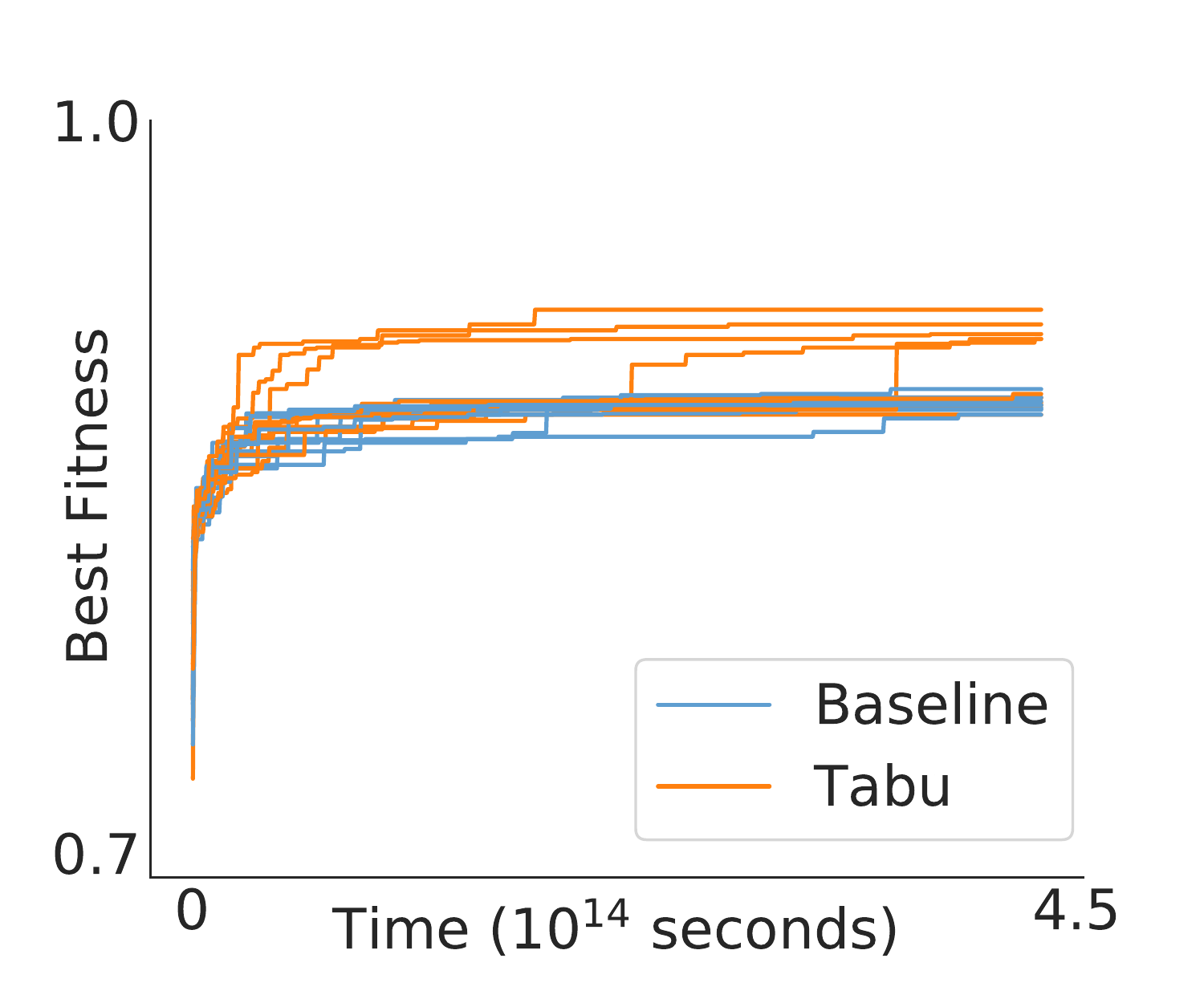}
        \caption{AutoML-Zero meta-training fitness time-courses. LEFT: FEC, MIDDLE: FCM, and RIGHT: tabulist.}
        \vspace{-2pt}
        \label{amlz_suppl_fig}
    \end{centering}
\end{figure}

}  

\supplsection{Time-course of AutoRL Experiments}{autorl_timecourses}{

Figure~\ref{autorl_suppl_fig} shows the time-course plots for the three techniques on the AutoRL setup.


\begin{figure}[!t]
    \begin{centering}
        \includegraphics[width=0.32\columnwidth]{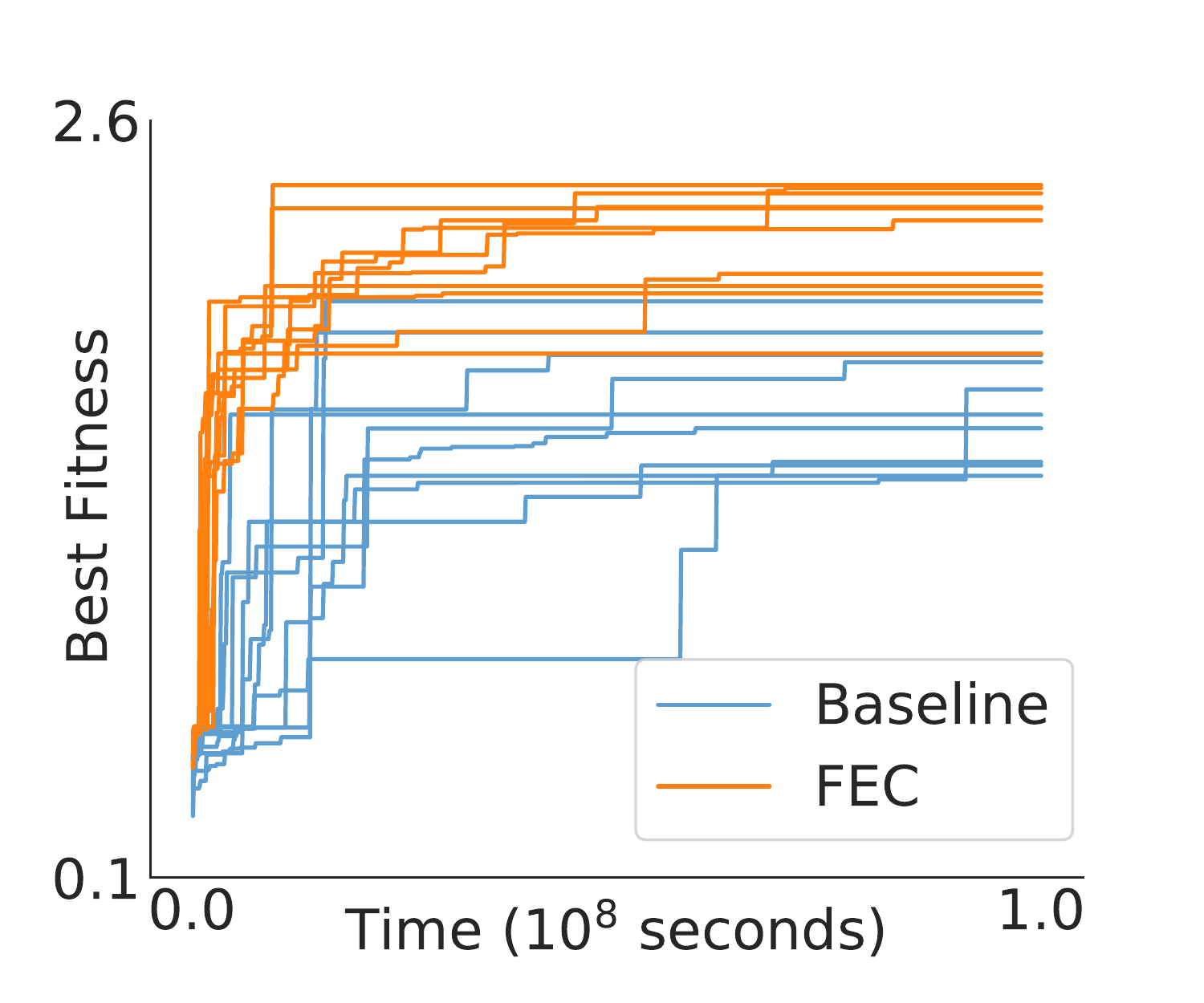}
        \includegraphics[width=0.32\columnwidth]{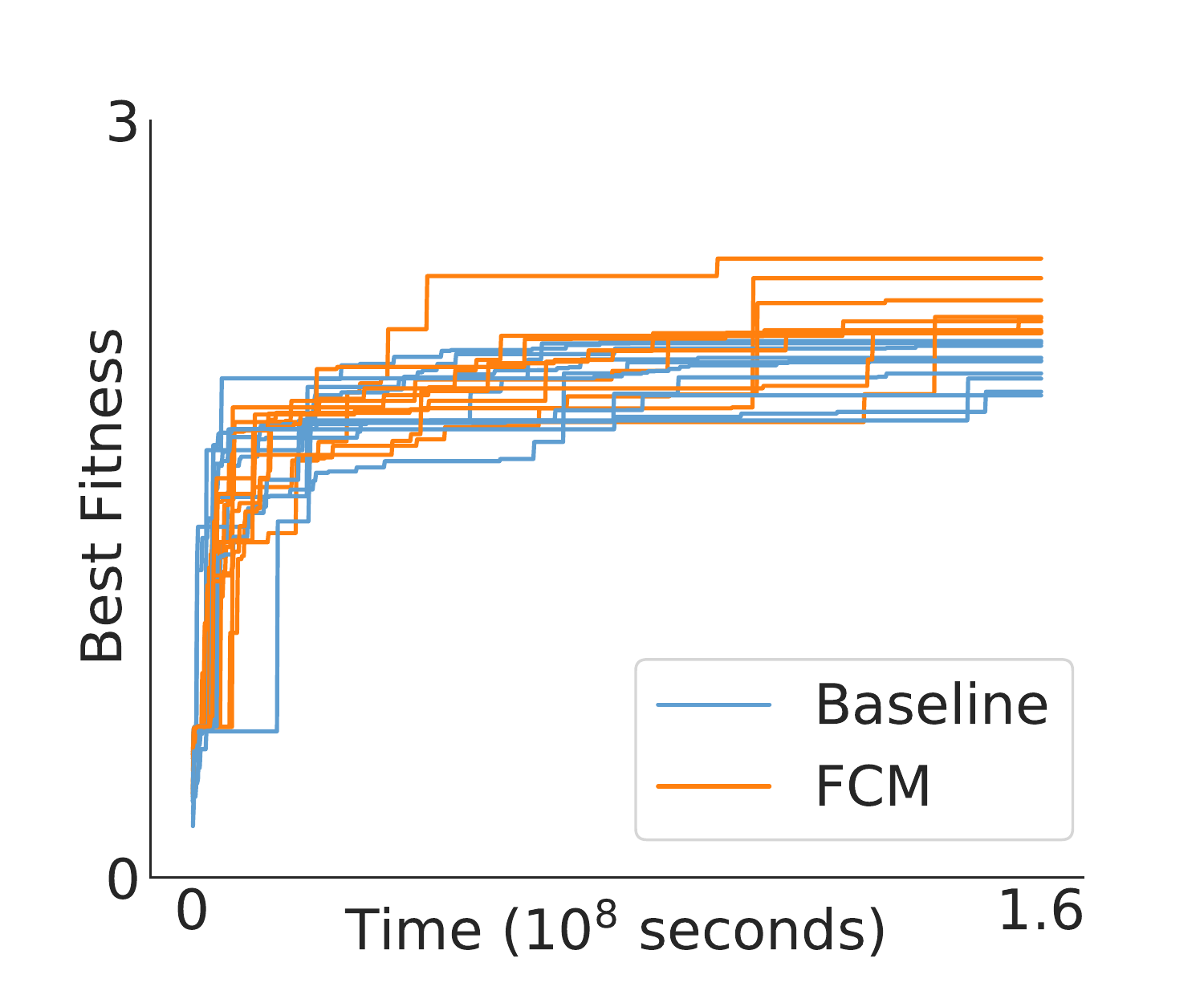}
        \includegraphics[width=0.32\columnwidth]{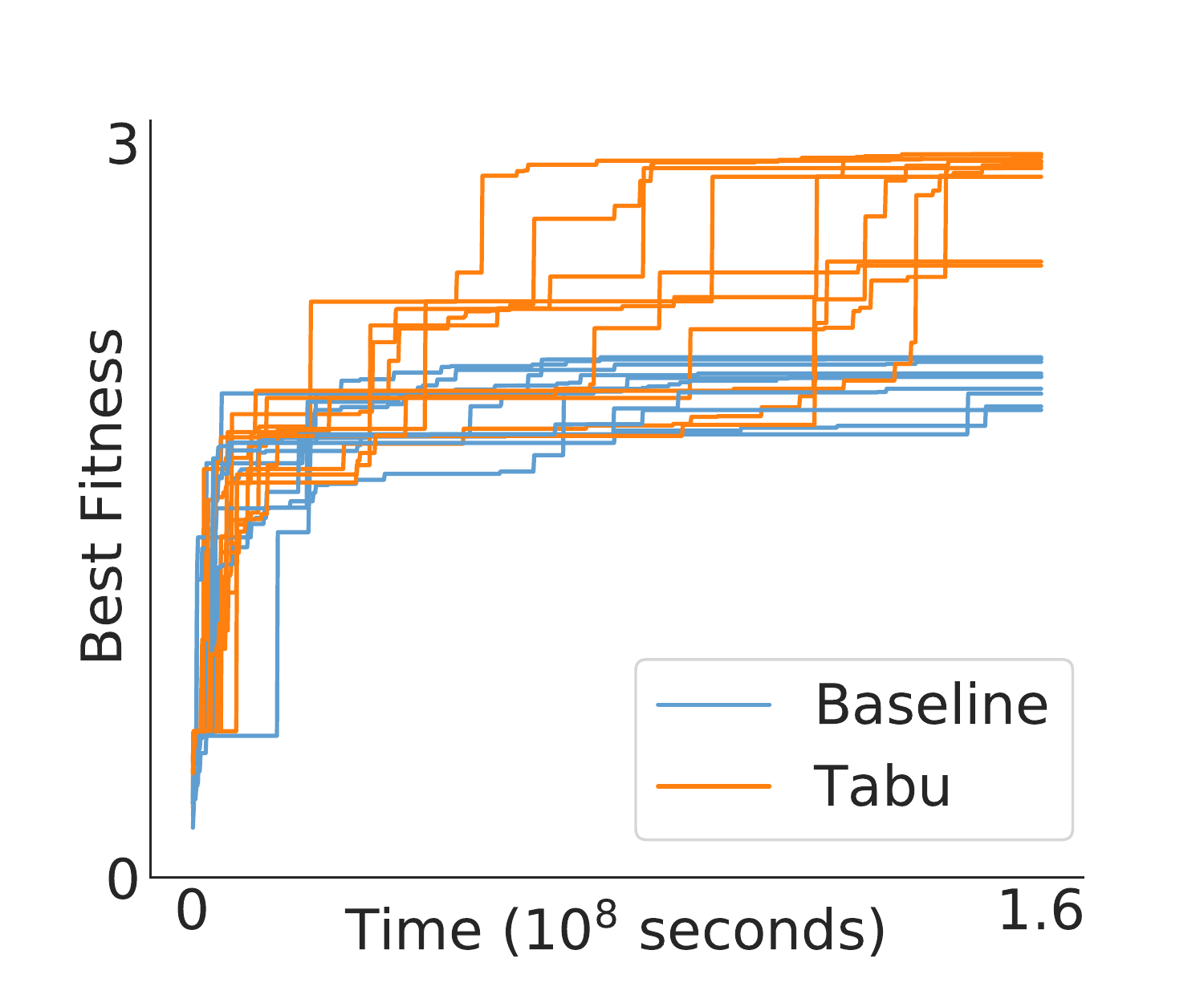}
        \caption{AutoRL meta-training fitness time-courses. LEFT: FEC, MIDDLE: FCM, and RIGHT: tabulist.}
        \vspace{-2pt}
        \label{autorl_suppl_fig}
    \end{centering}
\end{figure}

}  

\supplsection{FEC Variants}{fec_variants}{

Method~\ref{alg:fec_forget} shows a variant of FEC that has a small probability to forget a cached fitness after a cache hit.

\begin{algorithm}
\caption{Regularized Evolution \alghlight{with forgetful FEC (\alghlightcolor)}}
\label{alg:fec_forget}
\small
\begin{algorithmic}
\STATE $population \gets$ empty queue
\STATE \alghlight{$cache \gets $ empty hash table} \algspacedcomment{Candidate hash to fitness.}
\STATE $n = 0$  \algspacedcomment{Number of sampled candidates.}
\STATE \algcomment{P is the population size (a meta-parameter).}
\WHILE{$|population| < P$}
    \STATE $seed \gets$ RandomCandidate()
    \color{\alghlightcolor}
    \STATE $key \gets $ UnifiedFunctionalHash($seed$)
    \IF{$key$ in $cache$}
        \STATE $seed.fitness \gets cache[key]$  \algspacedcomment{Skips evaluation.}
    \STATE \algcomment{F is probability to forget key (a meta-parameter).}
    \IF{random.uniform < $F$}
        \STATE $cache.pop(key)$ \algspacedcomment{Forgets key.}
    \ENDIF
    \ELSE
    \color{black}
        \STATE $seed.fitness \gets$ Evaluate($seed$)
    \color{\alghlightcolor}
        \STATE $cache[key] \gets seed.fitness$
    \ENDIF
    \color{black}
    \STATE $population.enqueue(seed)$
    \STATE $n = n + 1$
\ENDWHILE
\STATE \algcomment{N is the total number of candidates (a meta-parameter).}
\WHILE{n < N}
    \STATE \algcomment{T is the tournament size (a meta-parameter).}
    \STATE $tournament \gets$ RandomSubset($population$, size=$T$)
    \STATE $parent \gets $ CandidateWithBestFitness($tournament$)
    \STATE $child \gets$ Mutate($parent$)
    
    \color{\alghlightcolor}
    \STATE $key \gets $ UnifiedFunctionalHash($child$)
    \IF{$key$ in $cache$}
        \STATE $child.fitness \gets cache[key]$  \algspacedcomment{Skips evaluation.}
    \STATE \algcomment{F is probability to forget key (a meta-parameter).}
    \IF{random.uniform < $F$}
        \STATE $cache.pop(key)$ \algspacedcomment{Forgets key.}
    \ENDIF
    \ELSE
    \color{black}
        \STATE $child.fitness \gets$ Evaluate($child$)
    \color{\alghlightcolor}
        \STATE $cache[key] \gets child.fitness$
    \ENDIF
    \color{black}
    \STATE $population.enqueue(child)$
    \STATE $population.dequeue()$ \algspacedcomment{Remove oldest.}
    \STATE $n = n + 1$
\ENDWHILE
\STATE \textbf{Return} CandidateWithBestFitness($population$)
\end{algorithmic}
\end{algorithm}

Method~\ref{alg:fea} shows a variant of FEC that aggregates fitnesses up to a maximum number of evaluations rather than evaluating a seed only once.

\begin{algorithm}
\caption{Regularized Evolution \alghlight{with aggregating FEC (\alghlightcolor)}}
\label{alg:fea}
\small
\begin{algorithmic}
\STATE $population \gets$ empty queue
\STATE \alghlight{$cache \gets $ empty hash table} \algspacedcomment{Candidate hash to fitness.}
\STATE $n = 0$  \algspacedcomment{Number of sampled candidates.}
\STATE \algcomment{P is the population size (a meta-parameter).}
\WHILE{$|population| < P$}
    \STATE $seed \gets$ RandomCandidate()
    \color{\alghlightcolor}
    \STATE $key \gets $ UnifiedFunctionalHash($seed$)
    \IF{$key$ in $cache$}
        \STATE \algcomment{M is max number of evaluations (a meta-parameter).}
        \IF{$cache[key].evals$ < $M$}
            \STATE $seed.fitness \gets$ Evaluate($seed$)
            \STATE $cache[key].fitness \gets$ UpdateAverage($seed.fitness$)
        \ELSE
            \STATE $seed.fitness \gets cache[key]$  \algspacedcomment{Skips evaluation.}
        \ENDIF
    \ELSE
    \color{black}
        \STATE $seed.fitness \gets$ Evaluate($seed$)
        \STATE $cache[key].fitness \gets$ UpdateAverage($seed.fitness$)
    \color{\alghlightcolor}
        \STATE $cache[key] \gets seed.fitness$
    \ENDIF
    \color{black}
    \STATE $population.enqueue(seed)$
    \STATE $n = n + 1$
\ENDWHILE
\STATE \algcomment{N is the total number of candidates (a meta-parameter).}
\WHILE{n < N}
    \STATE \algcomment{T is the tournament size (a meta-parameter).}
    \STATE $tournament \gets$ RandomSubset($population$, size=$T$)
    \STATE $parent \gets $ CandidateWithBestFitness($tournament$)
    \STATE $child \gets$ Mutate($parent$)
    
    \color{\alghlightcolor}
    \STATE $key \gets $ UnifiedFunctionalHash($child$)
    \IF{$key$ in $cache$}
        \STATE \algcomment{M is max number of evaluations (a meta-parameter).}
        \IF{$cache[key].evals$ < $M$}
            \STATE $seed.fitness \gets$ Evaluate($seed$)
            \STATE $cache[key].fitness \gets$ UpdateAverage($seed.fitness$)
        \ELSE
            \STATE $seed.fitness \gets cache[key]$  \algspacedcomment{Skips evaluation.}
        \ENDIF
    \ELSE
    \color{black}
        \STATE $seed.fitness \gets$ Evaluate($seed$)
        \STATE $cache[key].fitness \gets$ UpdateAverage($seed.fitness$)
    \color{\alghlightcolor}
        \STATE $cache[key] \gets seed.fitness$
    \ENDIF
    \color{black}
    \STATE $population.enqueue(child)$
    \STATE $population.dequeue()$ \algspacedcomment{Remove oldest.}
    \STATE $n = n + 1$
\ENDWHILE
\STATE \textbf{Return} CandidateWithBestFitness($population$)
\end{algorithmic}
\end{algorithm}

}  

\end{document}